\documentclass[twocolumn]{article}
\usepackage{arxiv}

\usepackage[utf8]{inputenc} 
\usepackage[T1]{fontenc}    
\usepackage{hyperref}       
\usepackage{url}            
\usepackage{booktabs}       
\usepackage{amsfonts}       
\usepackage{nicefrac}       
\usepackage{microtype}      %
\usepackage[pdftex]{graphicx}
\usepackage{amsmath}

\begin{document}

\title{Multi-Domain Level Generation and Blending with Sketches via Example-Driven BSP and Variational Autoencoders}

\author{
  Sam Snodgrass \\
  modl.ai\\
  Copenhagen, Denmark\\
  \texttt{sam@modl.ai} \\
   \And
 Anurag Sarkar \\
  Northeastern University\\
  Boston, MA, USA\\
  \texttt{sarkar.an@northeastern.edu}
  }



\twocolumn[
  \begin{@twocolumnfalse}

\maketitle
\begin{abstract}
Procedural content generation via machine learning (PCGML) has demonstrated its usefulness as a content and game creation approach, and has been shown to be able to support human creativity. An important facet of creativity is combinational creativity or the recombination, adaptation, and reuse of ideas and concepts between and across domains. In this paper, we present a PCGML approach for level generation that is able to recombine, adapt, and reuse structural patterns from several domains to approximate unseen domains. We extend prior work involving example-driven Binary Space Partitioning for recombining and reusing patterns in multiple domains, and incorporate Variational Autoencoders (VAEs) for generating unseen structures. We evaluate our approach by blending across $7$ domains and subsets of those domains. We show that our approach is able to blend domains together while retaining structural components. Additionally, by using different groups of training domains our approach is able to generate both 1) levels that reproduce and capture features of a target domain, and 2) levels that have vastly different properties from the input domain. 
\end{abstract}
\end{@twocolumnfalse}
]
\section{Introduction}
Procedural content generation via machine learning (PCGML)~\cite{summerville2017procedural} denotes a subgroup of PCG techniques that learn models of the type of content to be generated and then sample from those models to create new instances of the content (e.g. learn from a set of example game levels and then generate new levels having characteristics and properties  of the example levels). Common challenges of PCGML approaches are the generalizability of trained models across domains and finding or creating the training data needed for a given domain. As such, most PCGML level generation approaches have only explored a handful of level domains (predominantly, \textit{Super Mario Bros.}~\cite{summerville2016mariostring,guzdial2016game,snodgrass2017learning}, \textit{Kid Icarus}~\cite{snodgrass2017learning,snodgrass2016approach,sarkar2018blending,sarkar2019blending}, and \textit{The Legend of Zelda}~\cite{summerville2015samplinghyrule}). 

Recent work has begun exploring ways of addressing the above challenges.  Some have explored methods for leveraging existing training data to build models that generalize across several domains. These methods either try to supplement a new domain's training data with examples from other domains~\cite{snodgrass2016approach}, build multiple models and blend them together~\cite{guzdial2018automated,sarkar2018blending}, or directly build a model trained on multiple domains~\cite{sarkar2019blending}. Such approaches are pushing the field towards more generally applicable PCGML techniques, and open the door for more creative PCGML~\cite{guzdial2018combinatorial}. We propose an approach to level blending that falls in the latter category. Our approach blends levels from different domains together by finding and leveraging structural similarities between domains.

We build on existing PCGML research by combining two methods for generating levels, variational autoencoders (VAEs) and example--driven binary space partitioning (EDBSP). We leverage these approaches to model and generate levels at two levels of abstraction: one abstraction layer captures the structural information of the levels, and the other captures the finer domain--specific details such as object, enemy, and item placements. We test and evaluate our proposed approach across $7$ platforming games, $3$ of which have not been used as training or test domains in prior PCGML research, to the best of our knowledge. 

The main contributions of this paper are:
\begin{enumerate}
    \item A new PCGML approach for domain blending that combines two previous techniques, VAEs for modeling and generating structural level layouts and EDBSP for filling in those generated layouts by blending details from various domains.
    \item A multi-domain evaluation of the proposed approach exploring a broader range of domains than previous work.
\end{enumerate}

\section{Related Work}
Procedural content generation via machine learning (PCGML)~\cite{summerville2017procedural} describes a family of approaches for PCG that first learn a model of a domain from a set of training examples and then use that learned model to generate new content. Much PCGML research has focused on building models of individual domains in order to create new content within the chosen domain. A variety of approaches have been explored in pursuit of this goal (e.g., LSTMs~\cite{summerville2016mariostring}, DBNs~\cite{guzdial2016game}, Markov Models~\cite{snodgrass2017learning,dahlskog2014linear}, GANs~\cite{volz2018evolving}, VAEs~\cite{thakkar2019autoencoder}), and each has shown its ability to generate levels within a chosen domain. However, these techniques are only applicable in the domains in which they are trained, and rely on the existence of training data from the target domain. For this work, among the above approaches, we chose to use VAEs. Prior work has demonstrated their potential for blending domains \cite{sarkar2019blending} by learning continuous, latent models of input domains. Additionally, unlike GANs, VAEs also learn the mapping from the input domain to the latent domain which may make it more suitable in a co-creative design context. This is particularly useful since we hope to develop our approach into a mixed-initiative tool in the future. Moreover, VAEs also offer potential for controllability in the form of conditional VAEs.

Recently there has been work exploring PCGML approaches for blending domains and domain transfer. Guzdial and Riedl~\cite{guzdial2016learning} proposed a level blending model that blended different level styles within a single domain. Our work differs from theirs in that ours aims to blend between multiple domains. Guzdial and Riedl~\cite{guzdial2018automated} have also proposed a method for blending and combining complete games via conceptual expansion on learned graph representations of games. Our work instead focuses on blending levels by finding structural similarities between training domains and an input level sketch. Snodgrass and Ontan{\'o}n~\cite{snodgrass2016approach} presented a domain transfer approach for supplementing one domain with translated levels from another domain by finding mappings between the representations. In our work, we instead define a uniform abstract representation across domains which we use for finding structural similarities. Sarkar and Cooper~\cite{sarkar2018blending} trained separate LSTMs on multiple domains, and created blended levels by switching between the trained models. While the abstract level generation stage of our approach is trained separately on different domains, our full resolution level generation stage which performs the blending need not be retrained.

In blending and generating levels by combining together parts of different domains, our work, like past work referred above \cite{guzdial2016game,guzdial2016learning,guzdial2018automated,guzdial2018combinatorial,sarkar2018blending, sarkar2019blending}, also falls under combinational creativity \cite{boden2004creative}, the branch of creativity where new ideas and concepts are generated by combining existing ones in novel ways. Such methods can help in producing and exploring new design domains via blending and combination, as we attempt to do in this work by blending existing platformer domains to create new ones.

The approaches that are most relevant to our proposed work are Sarkar et al.'s~\cite{sarkar2019blending} use of VAEs for level generation and blending and Snodgrass'~\cite{snodgrass2019levels} example-driven BSP approach for generating levels from an input sketch. We present a hybrid model that combines these methods into a single pipeline allowing for the creation of new sketches by sampling from VAEs to create structural level sketches, and generating fully realized blended levels by using EDBSP with access to multiple domains to fill in the details of the sketches. This work extends previous EDBSP work by using multiple domains, allowing for domain blending; and by using sketches generated by VAEs, thus highlighting the versatility of the EDBSP approach.

\section{Methods}
At a high level, our proposed approach is composed of two stages. First, we use a variational autoencoder (VAE) to model and generate the abstracted structural patterns from a set of training levels in a given domain. Next, we pass a generated structural level sketch to an example-driven extension to the binary space partitioning algorithm. This algorithm generates a fully realized level by finding matching structural patterns in a set of training levels across multiple domains, and using those level sections to fill in the details resulting in a blended level. Below we describe how we represent our levels, and each stage of our approach in more detail.

\begin{figure*}[t]
\centering
\includegraphics[width=.8\textwidth]{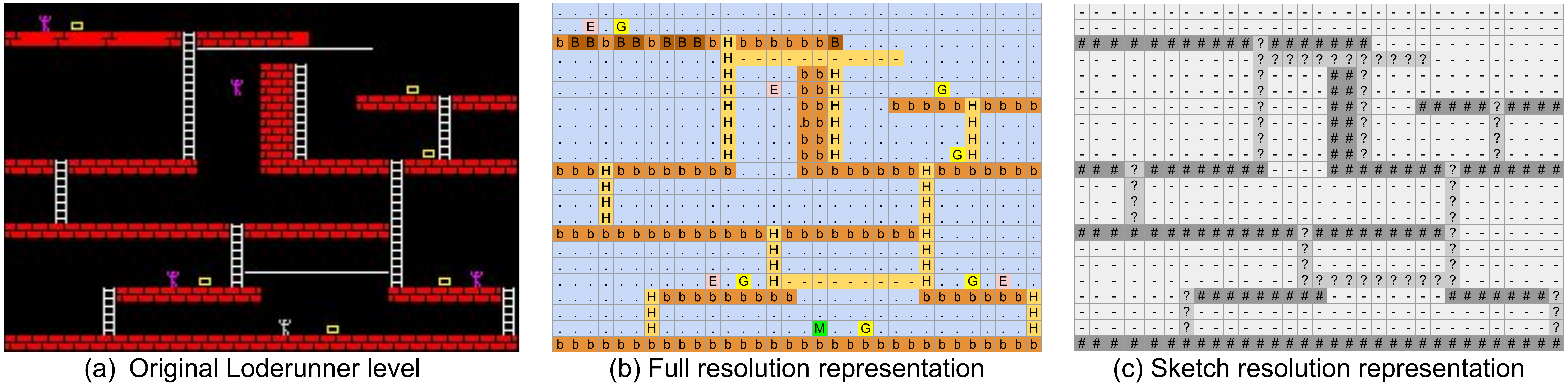}
\caption{This figure shows a \textit{Lode Runner} level (a), that same level represented with the full resolution representation (b), and that level represented with the sketch resolution representation (c).}\label{fig:levelRep}
\end{figure*}
\subsection{Level Representations} 
We demonstrate our approach using a set of NES platforming games (described in Section \ref{sec:dom}). We represent game levels with a tile grid where a cell can take a value from a set of tile types corresponding to elements of the domain. Figure \ref{fig:levelRep} (a-b) shows an example of such a representation. This style of representation is commonly used in PCGML approaches~\cite{summerville2017procedural} and is also used by the Video Game Level Corpus (VGLC)~\cite{summerville2016vglc}. Using this tile-based representation, we represent levels at two layers of abstraction, a \textbf{Full Resolution} layer and a \textbf{Sketch Resolution} layer. The tile types composing the full resolution layer differ between domains and correspond to specific structural components, interactive elements, enemies, and items in that domain. The sketch resolution layer, however, consists of the same three tile types across all domains: 
\begin{enumerate}
\item \textbf{\#}, representing a solid/impassable element; \item \textbf{-}, representing empty space or otherwise passable elements;
\item \textbf{?}, representing a wildcard that can be interpreted as either solid or empty. 
\end{enumerate}

\noindent The wildcard tile extends the previous sketch resolution representation~\cite{snodgrass2019levels}, and was included in this work to more easily capture structures that are not clearly represented by the empty or solid types (e.g., ladders). Figure \ref{fig:levelRep} shows a \textit{Lode Runner} level represented in these two abstractions. 



\subsection{Generating Sketch Resolution Levels}
Variational autoencoders (VAEs) \cite{kingma2013autoencoding} are generative models that learn continuous, latent representations of training data which can then be sampled to produce novel outputs. Such models consist of an encoder which maps the input data to a latent space and a decoder which maps from points in this latent space to outputs. While vanilla autoencoders \cite{hinton2006reducing} learn lower-dimensional latent representations of training data by only minimizing reconstruction error, VAEs additionally enforce the learned latent representation to model a continuous, probability distribution by minimizing the KL divergence between the latent distribution and a known prior (usually a Gaussian). Thus, similar to GANs, VAEs can generate novel variations of the training data in addition to being able to perform reconstruction. In this work, we used VAEs to generate levels at the sketch resolution layer, training a separate generative model for each domain.

\subsection{Generating Full Resolution Levels}\label{sec:BSP}
\begin{figure}
    \centering
    \includegraphics[width=\columnwidth]{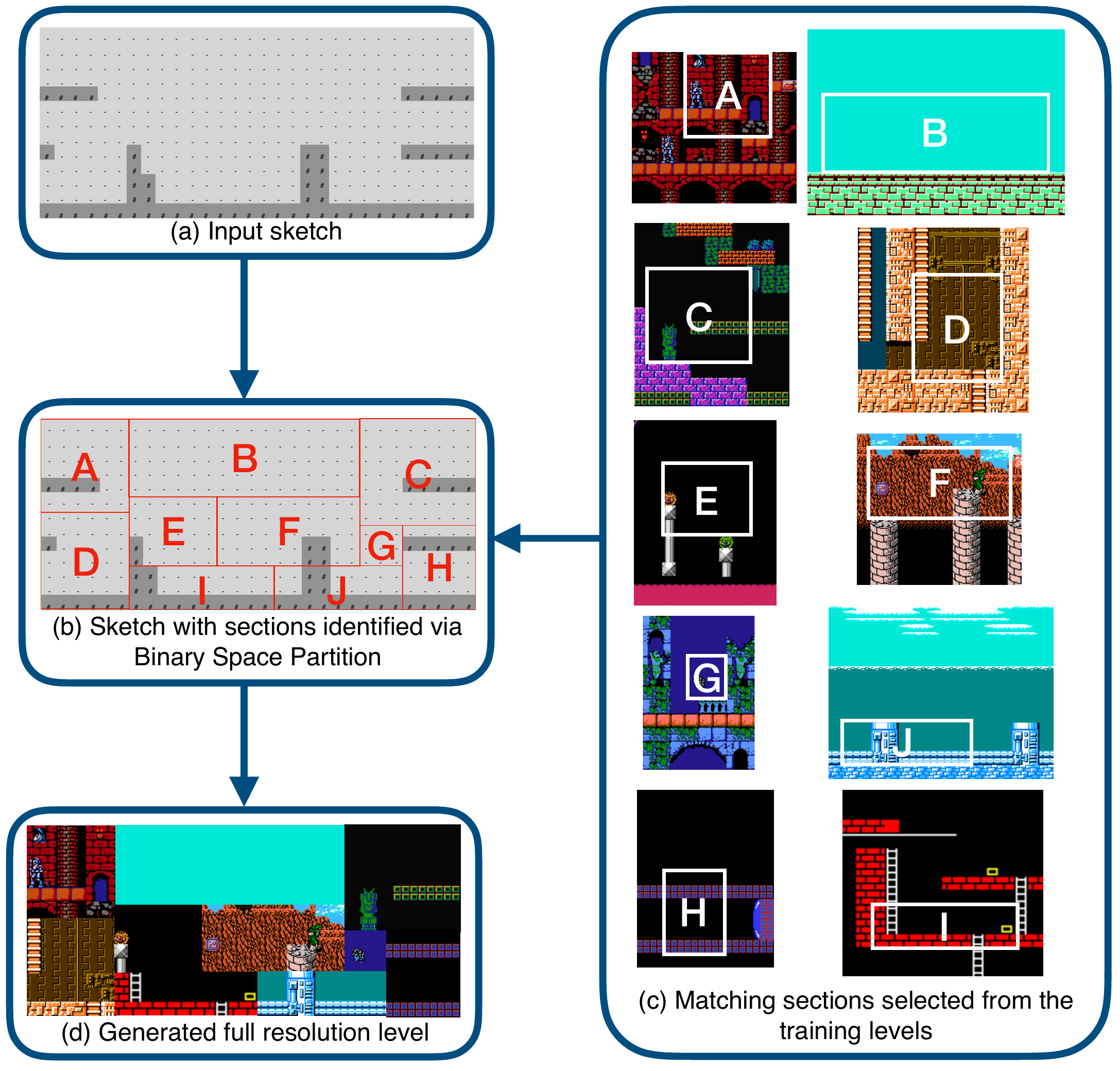}
    \caption{This figure shows the basic pipeline of the EDBSP algorithm. First, an input sketch is provided (a). This sketch can be chosen from the training data (as in Section \ref{sec:EDBSP-TS}) or generated with a VAE (as in Sections \ref{sec:VAE-GS} and \ref{sec:EDBSP-GS}).  Next, BSP is used to split the sketch into regions (b). Finally, structural matches for those sketch regions are found in the training data (c), and are used to create a full-resolution level (d).}
    \label{fig:EDBSP}
\end{figure}

Binary Space Partition (BSP)~\cite{togelius2016introduction} is a partitioning algorithm classically used in PCG for dungeon generation. The standard BSP algorithm recursively splits regions of a map into two smaller regions using a random orientation (vertical or horizontal) and positioning within the region until some end condition is met (e.g., a specified number of regions are created). Another process then takes those regions and converts them into a level (e.g., connects regions with doors, places enemies and keys, etc.). We use an extension of BSP called Example-driven Binary Space Partition (EDBSP)~\cite{snodgrass2019levels} which uses training data to fill in the details of the produced regions. Specifically, EDBSP is given an input sketch for a level (Figure \ref{fig:EDBSP}.a), and a set of training levels represented in both sketch and full resolution. BSP is then used to split the input sketch into regions (Figure \ref{fig:EDBSP}.b). For each region in the sketch, all the matching sketch resolution regions in the training levels are found, and one is chosen randomly from the set for that region (Figure \ref{fig:EDBSP}.c). The corresponding full resolution regions from the training set are then stitched together to produce the full resolution generated level (Figure \ref{fig:EDBSP}.d).

\section{Experiments}
\subsection{Domains}\label{sec:dom}
We test our level blending approach across seven domains chosen from NES platforming games: \textit{Castlevania (CV)} \cite{castlevania}, \textit{Kid Icarus (KI)} \cite{kidicarus}, \textit{Lode Runner (LR)} \cite{loderunner}, \textit{Mega Man (MM)} \cite{megaman}, \textit{Metroid (MT)} \cite{metroid}, \textit{Ninja Gaiden (NG)} \cite{ninjagaiden}, and \textit{Super Mario Bros. (SM)} \cite{supermario:nes}. Each of these domains differs from the others in the number of levels available and the size and shape of those levels (e.g., \textit{LR} has $150$ levels in the VGLC and \textit{KI} has $6$). This results in imbalanced data sets, which could lead to one domain being over represented in the generated levels simply by having more examples to draw from. To better investigate the relationships between the domains and the capabilities of our approach, we standardize the amount of training data from each domain. Specifically, we use a subset of levels from each domain such that training data from each domain is composed of approximately $18,000$ tiles\footnote{The set of training levels used in each domain can be found here: \url{https://bitbucket.org/FDG2020-Sketch/level-data/}}. Note, this value was chosen as it is the smallest number of tiles in our domains when using all data (i.e., the sum of tiles in all the \textit{CV} levels is $17,728$).

We divide our domains according to the presence of wildcards:

\begin{itemize}
    \item \textbf{WildCards (\textit{WC})}: This set contains domains with wildcard tiles in their sketch representations. This set includes \textit{CV} ($6$ levels), \textit{LR} ($25$ levels), \textit{MM} ($4$ levels),  and \textit{NG} ($8$ levels).
    \item \textbf{No WildCards ($\neg$\textit{WC})}: This set contains the domains that do not have wildcard tiles in their sketch representations. This set includes \textit{KI} ($5$ levels), \textit{MT} ($1$ section of the map split according to locked doors), and \textit{SM} ($6$ levels).
    \item \textbf{All Domains (\textit{ALL})}: This set is the union of the above sets.
\end{itemize}

\noindent 
\subsection{Experimental Setup}
We test our proposed approach on its ability to generate sketches and full resolution levels. We evaluate each of the stages of our approach individually, and then the full pipeline.

\subsubsection{Sketch Generation}\label{sec:VAE}
To test the sketch generation stage of our approach on its own, we trained a separate VAE on each of the domains, using the same overall architecture for each domain except for the dimensions of the input and output segments which we varied to suit each individual domain. For each VAE, the encoder consisted of 2 strided convolutional layers with batch normalization and leaky ReLU activation while the decoder consisted of 3 convolutional layers which were strided or non-strided as required by the dimensions of the specific domain. The decoder also used batch normalization but with ReLU activation. All models used a 32-dimensional latent space and were trained for 5000 epochs using the Adam optimizer and a learning rate of 0.001. For generation, we selected the model from the epoch which best minimized reconstruction error. All models were implemented using PyTorch \cite{paszke2017automatic}. Note that we use fixed-size windows instead of full levels for training and generation. This is to account for the variation in level sizes both across and within domains and for the fact that convolutional generative models work with fixed-size inputs and outputs. Thus, like prior work using such models for level generation \cite{volz2018evolving,sarkar2019blending}, we generated our training data by sliding a fixed-size window across the levels in each domain and trained our models using those segments obtained after filtering out ones that contained any empty space. We used the following dimensions for each domain:
\begin{itemize}
    \item \textit{CV}: 11x16
    \item \textit{KI}: 16x16
    \item \textit{MM}: 15x16
    \item \textit{SM}: 14x14
    \item \textit{LR}: 11x16
    \item \textit{MT}: 15x16
    \item \textit{NG}: 11x16
\end{itemize}
\noindent Note, we use different dimensions for the domains based on the height and width of the training levels. 

For each domain, we then generated $100$ sketch resolution sections of the fixed-size for that domain. For evaluating these sections, we computed the following metrics for each segment:

\begin{itemize}
    \item \textit{Density}: the proportion of solid tiles in a region.
    \item \textit{Non--Linearity}: how well a segment's topology fits to a line. It is the mean squared error of running linear regression on the highest point of each of the columns in a segment. A zero value indicates perfectly linear topology.
    \item \textit{Plagiarism}: a pairwise metric which counts the number of rows and columns a segment shares with another segment.
    \item \textit{E--Distance}: a measure of the distance between two distributions introduced by \cite{szekely2013energy} and suggested as a suitable metric for evaluating generative models by \cite{summerville2018expanding} due to certain desirable properties. The lower the E-distance, the more similar are the distributions being compared. For our evaluations, we computed E-distance using the \textit{Density} and \textit{Non-Linearity} of each of the 100 generated segments and that of a random sampling of 100 training segments, per domain.
\end{itemize}

\noindent Notice that we also computed these metrics for the training levels in order to compare against the generated set. The \textit{density}, \textit{non--linearity}, and \textit{E--distance} metrics measure how well the VAE can capture and replicate the structural patterns from the training levels. The \textit{plagiarism} metric measures how much the VAE copies from the training domain, and gives insight into whether the model is able to generate new sections or just replicate existing ones. Additionally, we computed \textit{self-plagiarism} i.e. how much pairs of training segments plagiarize from each other, as a means of understanding how well or poorly the plagiarism detected in the generative model compares with that which already exists in the training data. Due to the large number of training segments compared to the 100 generated segments per domain, for our evaluations, we computed plagiarism and self-plagiarism values using a random sampling of 100 training segments. Additionally, statistical comparisons between generated and training segments were also performed using this sampling.

\subsubsection{Conditional Sketch Generation} In addition to training a standard VAE on each sketch domain, we also trained a conditional VAE (CVAE)~\cite{sohn2015learning,yan2015conditional} on sketches from all domains taken together, with each sketch labeled with its corresponding domain. Conditional generative models~\cite{mirza2014conditional}, as the name suggests, enable generation of outputs conditioned on some given input. Such models are trained simply by concatenating training data instances with the data to be used for conditioning such as a class label, for example. Thus a CVAE trained as described above could enable generating sketches of a desired domain allowing for greater control in the generation process. For our CVAE, we used a different architecture than the regular VAEs described above, with the encoder and decoder both consisting of 2 linear layers, though the latent space was still 32-dimensional. The conditioning input was a one-hot encoded vector indicating the domain of the corresponding input sketch. For training, we used segments of dimension 11x16 for all domains as this was the largest window size that could accommodate all domains. The 11x16 segments were flattened to a single-dimensional input vector for the linear layers. Unfortunately, we did not obtain strong results using this approach and did not use CVAE-generated sketches as inputs to EDBSP for full level generation. However, conditioning the generation process still resulted in interesting outputs and opens up directions to consider for future work.

\subsubsection{Full Resolution Generation}
To test the full resolution generation stage of our approach on its own, we used each of the domains separately as input sketches to the EDBSP algorithm paired with different subsets of domains as the levels used for filling in the details and blending. For this, we chose a domain, then generated a total of $100$ full resolution levels for that domain divided evenly amongst the sketches (e.g., \textit{LR} has $25$ sketches, and therefore EDBSP generates four full resolution levels for each sketch; \textit{SM} has $6$ sketches, and EDBSP generates $16-17$ full resolution levels for each sketch). We perform this process for each domain, using each defined subset of domains (i.e., \textit{WC}, $\neg$\textit{WC}, \textit{ALL}) as the example full resolution levels to EDBSP. While using a given domain for its sketches, we removed it from its respective training data subset. This resulted in $300$ generated levels per domain, $100$ for each subset of domains. 

To test our full pipeline for level blending and generation (i.e., the full resolution generation stage combined with the sketch generation stage), we follow a similar procedure as above. We use the sketch sections generated for each domain using the VAE described in Section \ref{sec:VAE} as input to the EDBSP algorithm. For each domain we generate $10$ full resolution sections from each of the $100$ generated sketches. We perform this process with each defined subset of domains (\textit{WC}, $\neg$\textit{WC}, \textit{ALL}) as example full resolution levels for EDBSP, while removing the current sketch domain from the subsets. This results in $3000$ full resolution sections for each domain, $1000$ for each defined subset of domains.

We evaluated the generator and generated levels by computing:

\begin{itemize}
    \item \textit{Domain Proportion}: the proportion of the generated level that was generated using a given domain. This is computed as $\frac{\text{tiles from a domain in the level}}{\text{total tiles in the level}}$.
    \item \textit{Element Distribution Similarity}: the distribution of common level elements in the generated level (i.e., empty space, solid objects, enemies, items, hazardous objects, and climbable objects). We compute the KL divergence~\cite{kullback1951information} between this distribution in the generated levels and the training levels. 
\end{itemize}

\noindent The \textit{domain proportion} measure gives insight into the biases of our generator and representation. It can also help us understand which domains are structurally similar to one another and which contain more diverse structures. The \textit{element distribution similarity} measures if the generator is able to approximate a domain using examples from other domains. KL divergence has been used by others to guide level generators~\cite{lucas2019tile,volz2018evolving} and we use it here to measure relatedness between generated levels and the target domain.

\begin{table*}[tbh]
    \centering
    \caption{Computed metrics for VAE generated level sections. $\dagger$ on the generated values indicates statistically significant differences between the generated sections and the training levels in terms of the corresponding metric (using Wilcoxon test with $p\leq0.05$). Metric values for generated sections that are not significantly different from those for training levels are preferred since they indicate that the learned distribution is not significantly different than the distribution of the training domain. Similarly, the lower the E-distance, the closer the learned distribution is to the training distribution.}
\resizebox{\textwidth}{!}{
    \begin{tabular}{c||cc|cc|cc|c}
     & \multicolumn{2}{c|}{\textbf{Density}} & \multicolumn{2}{c|}{\textbf{Non--Linearity}} &\multicolumn{2}{c|}{\textbf{Plagiarism}} & \textbf{E--Distance} \\\hline
     \textbf{Domain} & \textit{Training} & \textit{Generated} & \textit{Training} & \textit{Generated} & \textit{Training} & \textit{Generated} & --\\ \hline\hline
     \textit{CV} & $18.71 \pm 11.02$ & $15.47 \pm 6.47$ & $3.55 \pm 3.16$ & $4.74 \pm 3.21^\dagger$ & $4.5 \pm 5.22$ & $4.32 \pm 3.34^\dagger$ & $1.63$ \\
     \textit{LR} & $36.61\pm16.39$ & $28.44\pm9.43^\dagger$ & $5.64\pm5.26$ & $6.23\pm3.62$ & $0.52\pm2.76$ & $0.26\pm0.69^\dagger$ & $2.80$ \\
     \textit{MM} & $41.25\pm14.74$ & $32.06\pm9.85^\dagger$ & $9.90\pm9.51$ & $16.53\pm10.73^\dagger$ & $2,18\pm4.41$ & $1.12\pm2.01^\dagger$ & $4.71$\\
     \textit{NG} & $18.07\pm10.83$ & $14.48\pm5.38$ & $4.18\pm4.35$ & $3.77\pm2.49$ & $4.96\pm4.10$ & $5.44\pm2.85^\dagger$ & $0.71$ \\
     \textit{KI} & $23.74\pm11.48$ & $15.69\pm5.41^\dagger$ & $12.28\pm11.77$ & $23.19\pm10.98^\dagger$ & $2.02\pm3.67$ & $1.61\pm1.76^\dagger$ & $9.72$ \\
     \textit{MT} & $43.36\pm13.07$ & $34.63\pm10.63^\dagger$ & $7.26\pm8.65$ & $14.74\pm10.49^\dagger$ & $1.67\pm3.56$ & $0.47\pm0.97$ & $9.03$ \\
     \textit{SM} & $10.81\pm4.74$ & $7.17\pm2.52^\dagger$ & $4.32\pm3.96$ & $2.59\pm2.96^\dagger$ & $15.16\pm5.45$ & $15.79\pm5.08^\dagger$ & $1.93$ \\

    \end{tabular}}
    \label{tab:vae}
\end{table*}
\section{Results and Discussion}
\begin{table}[tbh]
    \centering
    \caption{E--distance between CVAE-generated sketches and VAE-generated sketches from the corresponding domain and between 100 random sketches from the corresponding training domain. E-distances between the respective VAEs and training domains are also given for comparison.}
\resizebox{\columnwidth}{!}{
    \begin{tabular}{c||c|c|c}
    \textbf{Domain} &  \textbf{CVAE vs VAE} & \textbf{CVAE vs Train} & \textbf{VAE vs Train}\\ \hline \hline
    \textit{CV} & $7.97$ & $7.81$ & $1.56$\\
    \textit{LR} & $0.99$ & $3.91$ & $2.86$\\
    \textit{MM} & $4.61$ & $10.14$ & $5.87$\\
    \textit{NG} & $7.11$ & $3.91$ & $0.79$\\
    \textit{KI} & $14.44$ & $2.22$ & $11.38$\\
    \textit{MT} & $2.78$ & $14.49$ & $7.13$\\
    \textit{SM} & $10.36$ & $4.39$ & $2.41$\\
    \end{tabular}}
    \label{tab:cvae}
\end{table}
\subsection{Sketch Generation using VAEs}\label{sec:VAE-GS}
Table \ref{tab:vae} depicts the results of our evaluations of the sketch sections generated using the VAEs. The results suggest that the VAE performs the best in learning the distribution of \textit{NG} as exhibited by it having the lowest \textit{E-distance}, followed by \textit{CV}, \textit{SM} and \textit{LR} with the models for \textit{MM} and especially \textit{MT} and \textit{KI} performing worse with respect to these metrics. Generated sketch sections for \textit{NG} were the only ones to not be significantly different from the training set in terms of both \textit{Density} and \textit{Nonlinearity}, with those for \textit{CV} and \textit{LR} being significantly different in terms of one of these while those for the more E-distant \textit{MM}, \textit{MT} and \textit{KI} being different in terms of both. The outlier here is \textit{SM} which has the third lowest \textit{E-distance} but is significantly different in terms of both metrics. One possible explanation is that while the sections have similar mean values for the metrics, the individual values for the metrics on the generated and training sections may be very different from one another. 
Overall, the VAEs seem to do better in domains with less dense level structures such as \textit{SM}, \textit{CV} and \textit{NG} as opposed to those with higher density like \textit{MM}, \textit{MT} and \textit{KI}. This makes sense as it requires the model to learn less complex structural elements. Note that we used the same architecture for each domain so it is likely that the denser domains could have been better learned using more complex models. In a similar vein, domains with more uneven in-segment topology (i.e. having highly non-linear segments) are more difficult to learn than those with more linear segments. Since we trained our generators using fixed-size segments rather than whole levels, global level structure did not impact how well the generators were able to learn the input distribution. \textit{CV}, \textit{MT}, \textit{MM}, \textit{NG} progress both horizontally and vertically, \textit{SM} and \textit{KI} progress only horizontally and only vertically respectively, but differences in VAE performance were not detected along these lines. Rather, as our results show, it is the more local segment-based properties in the training sketches that influence the quality of generated sketches. To better depict the capabilities of the generators, as per the recommendations of \cite{summerville2018expanding}, for each domain, we show pairs of training and generated segments that were nearest and furthest with respect to each metric in Figure \ref{fig:VAEcomp} in the Appendix.

\begin{figure*}
    \centering
    \begin{tabular}{ccccccc}
    \includegraphics[width=.11\textwidth]{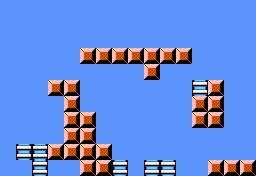}&
    \includegraphics[width=.11\textwidth]{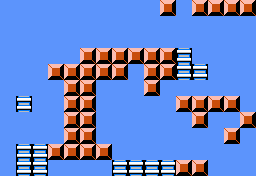}&
    \includegraphics[width=.11\textwidth]{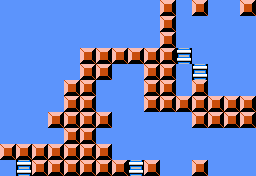}&
    \includegraphics[width=.11\textwidth]{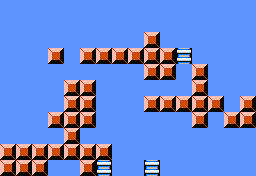}&
    \includegraphics[width=.11\textwidth]{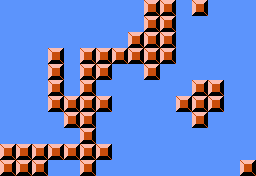}&
    \includegraphics[width=.11\textwidth]{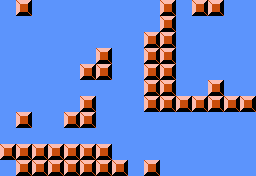}&
    \includegraphics[width=.11\textwidth]{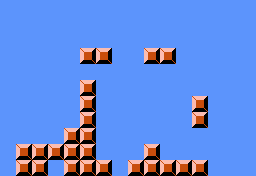}\\
    (a) \textit{CV} & (b) \textit{LR} &(c) \textit{MM} &(d) \textit{NG} &(e) \textit{KI} &(f) \textit{MT} &(g) \textit{SM}\\
        \end{tabular}
    \caption{Sketch sections generated by the CVAE using the same input vector but different domain as the conditioning input}
    \label{fig:CVAE}
\end{figure*}

As stated previously, our conditional sketch generation efforts did not produce strong results with most metrics being very different from the input domains with statistical significance. Table \ref{tab:cvae} shows E-distances between 100 CVAE-generated sketches vs. 100 sketches generated using the VAE of that corresponding domain and vs. 100 sketches sampled randomly from the training set. All distances are higher than those between the VAE-generated sketches and the training domain, with the exception of \textit{KI} which proved to have the lowest E-distance for the CVAE case while having the highest for the VAE. However, seeing how the CVAE does not do well in other domains, we attribute this to be circumstantial and leave further investigations into the CVAE for future work. As exemplars of what is possible using this approach, Figure \ref{fig:CVAE} shows segments generated using the same random vector conditioned on different domains.

\begin{table*}[tbh]
\caption{Distribution of domain proportions in full resolution levels generated from existing sketches}
\begin{center}
\resizebox{\textwidth}{!}{
\begin{tabular}{c||ccccccc}
        & \multicolumn{7}{c}{\textbf{ALL}}\\ \hline\hline
    \textbf{Input} & \textit{CV} & \textit{LR} &\textit{MM} & \textit{NG} & \textit{KI} & \textit{MT} & \textit{SM} \\ \hline
    \textit{CV} & --  & $\mathbf{0.263 \pm 0.066}$  & $0.148 \pm 0.025$  & $0.163 \pm 0.028$  & $0.110 \pm 0.026$  & $0.150 \pm 0.030$  & $0.166 \pm 0.035$  \\
    \textit{LR} & $0.156 \pm 0.102$  & --  & $\mathbf{0.270 \pm 0.120}$  & $0.154 \pm 0.093$  & $0.109 \pm 0.079$  & $0.164 \pm 0.109$  & $0.146 \pm 0.138$  \\
\textit{MM} & $0.157 \pm 0.025$  & $\mathbf{0.323 \pm 0.063}$  & --  & $0.176 \pm 0.020$  & $0.146 \pm 0.028$  & $0.049 \pm 0.105$  & $0.148 \pm 0.020$  \\
    \textit{NG} & $0.150 \pm 0.064$  & $\mathbf{0.358 \pm 0.140}$  & $0.134 \pm 0.056$  & --  & $0.078 \pm 0.043$  & $0.134 \pm 0.053$  & $0.146 \pm 0.062$  \\
    \textit{KI} & $0.113 \pm 0.034$  & $\mathbf{0.394 \pm 0.128}$  & $0.141 \pm 0.029$  & $0.119 \pm 0.032$  & --  & $0.127 \pm 0.027$  & $0.106 \pm 0.036$  \\ 
    \textit{MT} & $0.123\pm 0.005$  & $0.254 \pm 0.008$  & $\mathbf{0.262 \pm 0.007}$  & $0.135 \pm 0.006$  & $0.125 \pm 0.006$  & --  & $0.102 \pm 0.005$  \\ 
    \textit{SM} & $0.171 \pm 0.036$  & $\mathbf{0.287 \pm 0.109}$  & $0.151 \pm 0.036$  & $0.172 \pm 0.047$  & $0.096 \pm 0.044$  & $0.123 \pm 0.033$  & --  \\ \hline 
    & \multicolumn{4}{c||}{\textbf{WC}} &\multicolumn{3}{c}{$\neg$\textbf{WC}}\\ \hline\hline
    \textbf{Input} & \textit{CV} & \textit{LR} &\textit{MM} &\multicolumn{1}{c||}{\textit{NG}} & \textit{KI} & \textit{MT} & \textit{SM} \\ \hline
    \textit{CV} & --  & $\mathbf{0.426 \pm 0.069}$  & $0.273 \pm 0.033$  &\multicolumn{1}{c||}{$0.301 \pm 0.052$}   & $0.293 \pm 0.042$  & $0.348 \pm 0.05$  & $\mathbf{0.359 \pm 0.057}$  \\
    \textit{LR} & $0.276 \pm 0.144$  & --  & $\mathbf{0.463 \pm 0.172}$  &\multicolumn{1}{c||}{$0.260 \pm 0.119$}   & $0.311 \pm 0.139$  & $\mathbf{0.367 \pm 0.168}$  & $0.323 \pm 0.189$  \\
    \textit{MM} & $0.249 \pm 0.025$  & $\mathbf{0.468 \pm 0.037}$  & --  &\multicolumn{1}{c||}{$0.283 \pm 0.024$}   & $0.263 \pm 0.023$  & $\mathbf{0.465 \pm 0.028}$  & $0.272 \pm 0.026$  \\
    \textit{NG} & $0.262 \pm 0.073$  & $\mathbf{0.496 \pm 0.123}$  & $0.242 \pm 0.088$  &\multicolumn{1}{c||}{--}   & $0.263 \pm 0.068$  & $0.354 \pm 0.077$  & $\mathbf{0.383 \pm 0.083}$  \\ \hline
    \textit{KI} & $0.151 \pm 0.053$  & $\mathbf{0.498 \pm 0.132}$  & $0.192 \pm 0.046$  &\multicolumn{1}{c||}{$0.160 \pm 0.045$}   & --  & $\mathbf{0.569 \pm 0.050}$  & $0.431 \pm 0.050$  \\ 
    \textit{MT} & $0.169 \pm 0.006$  & $0.321 \pm 0.009$  & $\mathbf{0.327 \pm 0.007}$  &\multicolumn{1}{c||}{$0.183 \pm 0.006$}   & $\mathbf{0.634 \pm 0.008}$  & --  & $0.366 \pm 0.008$  \\ 
    \textit{SM} & $0.227 \pm 0.046$ & $\mathbf{0.346 \pm 0.088}$ & $0.193 \pm 0.032$  &\multicolumn{1}{c||}{$0.234 \pm 0.050$}   & $0.435 \pm 0.093$  & $\mathbf{0.565 \pm 0.093}$   & --  \\ 
\end{tabular}}
\end{center}
\label{tbl:dstr}
\end{table*}

\begin{table*}[tbh]
\caption{Distribution of domain proportions in full resolution levels generated from generated sketches}
\begin{center}
\resizebox{\textwidth}{!}{
\begin{tabular}{c||ccccccc}
        & \multicolumn{7}{c}{\textbf{ALL}}\\ \hline\hline
    \textbf{Input} & \textit{CV} & \textit{LR} &\textit{MM} & \textit{NG} & \textit{KI} & \textit{MT} & \textit{SM} \\ \hline
    \textit{CV} & -- & $\mathbf{0.652\pm0.235}$ & $0.105\pm0.143$ & $0.080\pm0.129$ & $0.032\pm0.678$ & $0.516\pm0.101$ & $0.080\pm0.130$ \\
    \textit{LR} & $0.148\pm0.143$ & -- & $\mathbf{0.280\pm0.191}$ & $0.192\pm0.169$ & $0.165\pm0.152$ & $0.086\pm0.109$ & $0.128\pm0.163$\\
    \textit{MM} & $0.053\pm0.078$ & $\mathbf{0.607\pm0.182}$ & -- & $0.087\pm0.100$ & $0.049\pm0.068$ & $0.133\pm0.0115$ & $0.070\pm0.098$\\
    \textit{NG} & $0.077\pm0.121$ & $\mathbf{0.598\pm0.248}$ & $0.112\pm0.170$ & -- &$0.029\pm0.065$ & $0.051\pm0.101$ & $0.133\pm0.173$ \\
    \textit{KI} & $0.033\pm$ & $\mathbf{0.814\pm0.122}$ & $0.055\pm0.070$ & $0.041\pm0.062$ & -- & $0.020\pm0.041$ & $0.037\pm0.060$\\
    \textit{MT} & $0.039\pm0.058$ & $\mathbf{0.646\pm0.160}$ & $0.183\pm0.132$ & $0.057\pm0.074$ & $0.047\pm0.059$ & -- & $0.028\pm0.051$\\
    \textit{SM} & $0.147\pm0.185$ & $\mathbf{0.485\pm0.281}$ & $0.116\pm0.179$ & $0.128\pm0.174$ & $0.021\pm0.052$ & $0.102\pm0.172$ & --\\\hline
    & \multicolumn{4}{c||}{\textbf{WC}} &\multicolumn{3}{c}{$\neg$\textbf{WC}}\\ \hline\hline
    \textbf{Input} & \textit{CV} & \textit{LR} &\textit{MM} &\multicolumn{1}{c||}{\textit{NG}} & \textit{KI} & \textit{MT} & \textit{SM} \\ \hline
    \textit{CV} & -- & $\mathbf{0.719\pm0.221}$ & $0.151\pm0.170$ & \multicolumn{1}{c||}{$0.120\pm0.161$} & $0.306\pm0.179$ & $0.270\pm0.197$ & $\mathbf{0.423\pm0.223}$\\
    \textit{LR} & $0.246\pm0.175$ & -- & $\mathbf{0.436\pm0.206}$ & \multicolumn{1}{c||}{$0.318\pm0.199$} & $\mathbf{0.443\pm0.197}$ & $0.254\pm0.168$  & $0.303\pm0.209$ \\
    \textit{MM} & $0.098\pm0.103$ & $\mathbf{0.753\pm0.153}$ & -- & \multicolumn{1}{c||}{$0.149\pm0.124$} & $0.316\pm0.143$ & $\mathbf{0.430\pm0.162}$ & $0.254\pm0.163$\\
    \textit{NG} & $0.134\pm0.166$ & $\mathbf{0.692\pm0.234}$ & $0.173\pm0.188$  & \multicolumn{1}{c||}{--} & $0.266\pm0.176$  & $0.251\pm0.182$ & $\mathbf{0.483\pm0.221}$\\\hline
    \textit{KI} & $0.042\pm0.063$ & $\mathbf{0.839\pm0.114}$ & $0.068\pm0.075$ & \multicolumn{1}{c||}{$0.052\pm0.065$} & -- & $0.480\pm0.138$ & $\mathbf{0.520\pm0.138}$\\
    \textit{MT} & $0.051\pm0.069$ & $\mathbf{0.688\pm0.149}$ & $0.195\pm0.127$ & \multicolumn{1}{c||}{$0.067\pm0.075$} & $\mathbf{0.689\pm0.131}$  & -- & $0.312\pm0.131$\\
    \textit{SM} & $0.161\pm0.187$ & $\mathbf{0.532\pm0.274}$ & $0.149\pm0.194$ & \multicolumn{1}{c||}{$0.158\pm0.190$} & $0.261\pm0.189$ & $\mathbf{0.739\pm0.189}$ & -- \\
\end{tabular}}
\end{center}
\label{tbl:dstrGen}
\end{table*}

\subsection{EDBSP with Training Sketches}\label{sec:EDBSP-TS}
\begin{figure}[h]
    \centering
    \includegraphics[width=.2\columnwidth]{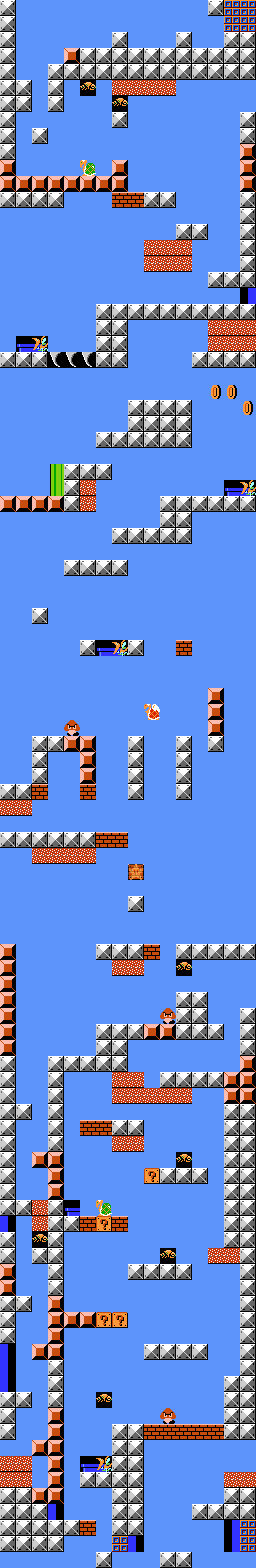}
    \includegraphics[width=.2\columnwidth]{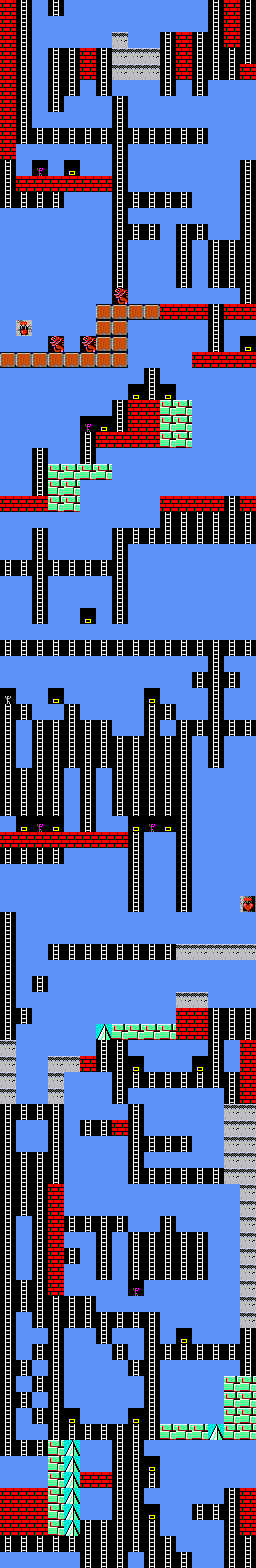}
    \caption{The generated \textit{KI} level with the lowest KL divergence ($0.038$) in the $\neg$\textit{WC} generated set (left); and the generated \textit{KI} level with the highest KL--divergence ($0.378$) in the \textit{WC} generated set (right). Both are cropped for space.}
    \label{fig:KIKL}
\end{figure}

Table \ref{tbl:dstr} shows the results of the \textit{domain proportion} for each domain, across sets of levels generated with existing sketches. What is immediately apparent is that \textit{Lode Runner} (\textit{LR}) dominates many of the generated levels when it is included in the example set, particularly in the \textit{WC} set where there are fewer example domains. This is likely because \textit{LR} levels have a large proportion of wildcard tiles in their sketches as compared to the other domains ($12\%$ of tiles in \textit{LR} while next highest is $~3\%$ of tiles in \textit{Mega Man} (\textit{MM})). Due to how EDBSP performs pattern matching with the wildcard tiles, this causes many more viable matches for \textit{LR} than for other domains, and an inflation of the prominence of \textit{LR} in the generated levels. An example of this is shown in a generated \textit{KI} level in Figure \ref{fig:KIKL} (right).

\begin{figure}[h]
    \centering
    \includegraphics[width=\columnwidth]{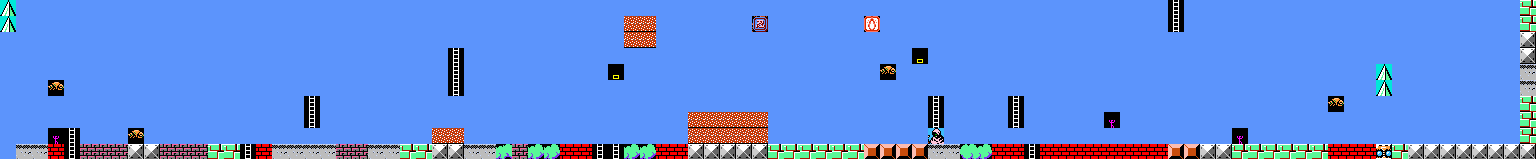}
    \includegraphics[width=\columnwidth]{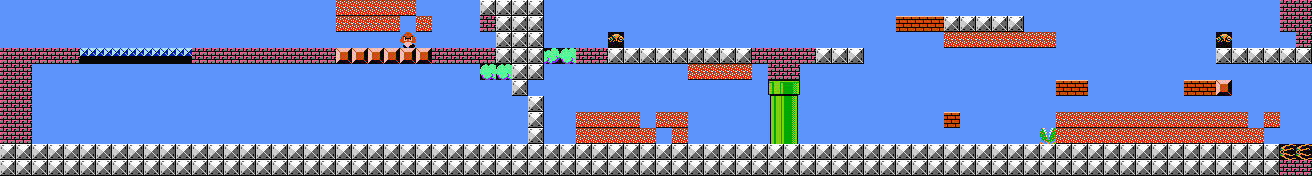}
    \caption{The generated \textit{CV} level with the lowest KL divergence ($0.044$) in the \textit{ALL} generated set (above); and the generated \textit{CV} level with the highest KL--divergence ($0.152$) in the $\neg$\textit{WC} generated set (below). Both are cropped for space.}
    \label{fig:CVKL}
\end{figure}

The only generated set which uses \textit{LR}, but does not have it as the most common domain is \textit{Metroid} (\textit{MT}), where \textit{MM} is the most common.  This may be due to the similarity in the structural layouts of \textit{MM} and \textit{MT} levels (i.e., both domains' levels consist of large sections of horizontal and vertical traversals with smaller obstacles mixed in). Additionally, as mentioned above, \textit{MM} has the second highest proportion of wildcard tiles. However, this relationship is not reciprocal. When using \textit{ALL} domains, \textit{MT} is the least frequent domain in the \textit{MM} levels. This shows that the wildcard tiles are important when finding matching examples in the training data, but when present in the input sketch they lead to more matches in all domains. When generating with the $\neg$\textit{WC} example set, we see that \textit{MT} is typically the most prevalent (or near to the most prevalent) domain, displaying the structural diversity of the domain. Lastly, while the generated blended levels in Figures \ref{fig:CVKL} and \ref{fig:KIKL} may not be playable using rules from the input sketch domain, we are not expecting the final levels to replicate a single domain and so playability within that domain is not required. Further, recall that the end goal is a mixed-initiative tool where the user could be controlling for final quality of the blended levels.

\begin{table}[tbh]
    \centering
    \caption{KL divergence between the training levels and levels generated from existing sketches using the distribution of game elements in the levels.}
    \resizebox{\columnwidth}{!}{
    \begin{tabular}{c||c|ccc}
    \textbf{Domain} &  \textbf{Uniform} & \textbf{ALL} & \textbf{WC} & $\neg$ \textbf{WC}\\ \hline \hline
    \textit{CV} & $1.098$ & $\mathbf{0.062}$ & $0.089$ & $0.136$\\
    \textit{LR} & $0.790$ & $0.280$ & $\mathbf{0.252}$ & $1.027$\\
    \textit{MM} & $0.930$ & $\mathbf{0.067}$ & $0.163$ & $0.276$\\
    \textit{NG} & $1.192$ & $\mathbf{0.087}$ & $0.119$ & $\mathbf{0.087}$\\\hline
    \textit{KI} & $1.195$ & $0.178$ & $0.208$ & $\mathbf{0.040}$\\
    \textit{MT} & $0.873$ & $0.098$ & $0.111$ & $\mathbf{0.081}$\\
    \textit{SM} & $1.374$ & $0.088$ & $0.105$ & $\mathbf{0.045}$\\
    \end{tabular}}
    \label{tab:tileKLdiv}
\end{table}
\begin{table}[tbh]
    \centering
    \caption{KL divergence between the training levels and levels generated from VAE-generated sketch sections using the distribution of game elements in the levels.}
\resizebox{\columnwidth}{!}{
    \begin{tabular}{c||c|ccc}
    \textbf{Domain} &  \textbf{Uniform} & \textbf{ALL} & \textbf{WC} & $\neg$ \textbf{WC}\\ \hline \hline
    \textit{CV} & $1.098$ & $0.248$ & $0.264$ & $\mathbf{0.124}$\\
    \textit{LR} & $0.790$ & $0.242$ & $\mathbf{0.222}$ & $1.151$\\
    \textit{MM} & $0.930$ & $\mathbf{0.196}$ & $0.345$ & $0.260$\\
    \textit{NG} & $1.192$ & $0.219$ & $0.254$ & $\mathbf{0.075}$\\\hline
    \textit{KI} & $1.195$ & $0.567$ & $0.581$ & $\mathbf{0.061}$\\
    \textit{MT} & $0.873$ & $0.415$ & $0.434$ & $\mathbf{0.146}$\\
    \textit{SM} & $1.374$ & $0.170$ & $0.182$ & $\mathbf{0.045}$\\
    \end{tabular}}
    \label{tab:tileKLdivGen}
\end{table}

\subsection{Training Sketches vs Generated Sketches}\label{sec:EDBSP-GS}
Table \ref{tab:tileKLdiv} shows the KL divergence between the \textit{element distributions} of the training levels, generated levels, and a uniform distribution. Here we can see the impact the choice of training data has on generating full resolution levels. Specifically, the last three rows containing the $\neg$\textit{WC} domains show that KL divergence is lowest when using the associated $\neg$\textit{WC} training set. Alternatively, in the first four rows, the \textit{WC} domains tend to have the lowest KL divergence with the levels generated using all the training data. This result shows that generally, the \textit{WC} domain sketches benefit from a variety of training data with different properties, while the $\neg$\textit{WC} domain sketches are best filled with details from more similar domains. The outlier in this table is again \textit{LR}, which has a higher KL divergence across all generated sets than any other domain, and when using the $\neg$\textit{WC} training domains, has a higher KL divergence than when compared with the uniform distribution of elements. This is due to the high frequency of special structures in \textit{LR}.

Table \ref{tbl:dstrGen} shows the distribution of the domains in the full resolution sections generated using VAE-generated sketch sections. In this table we can see the same trends as when generating with existing sketches, but more exaggerated. Specifically, we see that \textit{LR} dominates the generated sections to a higher degree. This likely results from the interaction between the size of the sections generated, and the way the partitioning algorithm divides the regions. EDBSP splits sections using the minimum dimension of the input as the maximum size of a region. In smaller areas, this can result in the large portions of the section being assigned one domain, which is likely to be assigned to \textit{LR} given its large number of wildcards. 

Table \ref{tab:tileKLdivGen} shows the KL divergence between \textit{element distributions} in the training levels, VAE-generated sketches, and a uniform distribution. This table reflects the disproportionate representation of \textit{LR} in the generated sections. Notably, the KL divergence has increased by large proportion in the \textit{ALL} and \textit{WC} generation sets, with much less variation in $\neg$\textit{WC} generated sets. Additionally, the lowest KL divergences are different from those in Table \ref{tab:tileKLdiv} for \textit{CV} and \textit{NG}, the \textit{WC} domains with lower wildcard proportions.

The results all point towards the importance of the choice of domains when blending. If approximating a specific domain or style of level is desired, then the domains with levels similar to the desired style should be chosen. For example, approximating \textit{KI} using \textit{MT} and \textit{SM} leads to similar \textit{element distributions}. On the other hand, if replicating a specific domain or style is not the goal, but instead exploration of new potential domains, then mixing a variety of different domains and examples can result in levels that have vastly different properties from the input domains. For example, blending \textit{MT}, \textit{KI}, and \textit{SM} with a sketch from \textit{MM} results in levels with \textit{element distributions} very different from the sketch domain.

\section{Conclusions}
We presented a novel, hybrid PCGML approach that combines the use of Example-Driven Binary Space Partitioning and VAEs to generate and blend levels using multiple domains. Our results demonstrate that different level generation and blending style goals (integrity vs. novelty, for example) can be traded off using different choices of domains. We consider several avenues for future work.

The experiments revealed that the choice of training domain representation can have a large impact on the resulting generated levels when blending. One avenue we would like to explore is intelligent automatic grouping of training domains. For example, if we know a priori that a set of domains has similar structures and game element distributions vs a set of domains that has similar structures but very different element distributions, we can better leverage the training data to guide the generator towards the users' goals (e.g., novelty vs replication).

Similarly, future work could also explore different choices of abstractions. In this work, the solid/empty sketch resolution abstraction allowed us to blend domains based on structural similarities but other abstractions could be defined based on other affordances such as those given in the Video Game Affordances Corpus \cite{bentley2019videogame}. Abstractions based on such affordances could potentially enable blending across different genres that do not share the same structural patterns and properties.

Our conditional sketch generation results were not optimal and conditioning a combined model failed to approximate the distributions of individual domains. It is likely the architecture was not well suited to the problem but even so, the results depicted in Figure \ref{fig:CVAE} suggest that this may be a promising direction to pursue. Successfully training such models would eliminate reliance on separate models for each domain for sketch generation. We would also like to explore other established blending and style transfer approaches. For example, how would CycleGAN~\cite{zhu_unpaired_2017} or pix2pix~\cite{isola_image_2017} perform on tile-resolution data instead of pixel resolution?

Lastly, we are interested in developing this approach into a mixed-initiative tool for level design and blending by allowing users to select their input domains, and create sketches for the EDBSP algorithm to fill in. By leveraging VAEs to generate new sketches, we have shown that the EDBSP approach is able to handle unseen sketches well, and therefore user generated sketches should be usable by the algorithm. Furthermore, the inner workings of the EDBSP algorithm are straightforward and explainable; and we would like to perform a user study to determine if that explainability increases usability in a mixed-initiative setting.


\section*{Appendix}\label{sec:app}

\begin{figure*}[h]
    \centering
    \begin{tabular}{ccc}
         \multicolumn{3}{c}{(a) Comparison of generated \textit{CV} sketch sections and training sketch sections}\\
         \includegraphics[width=0.28\textwidth]{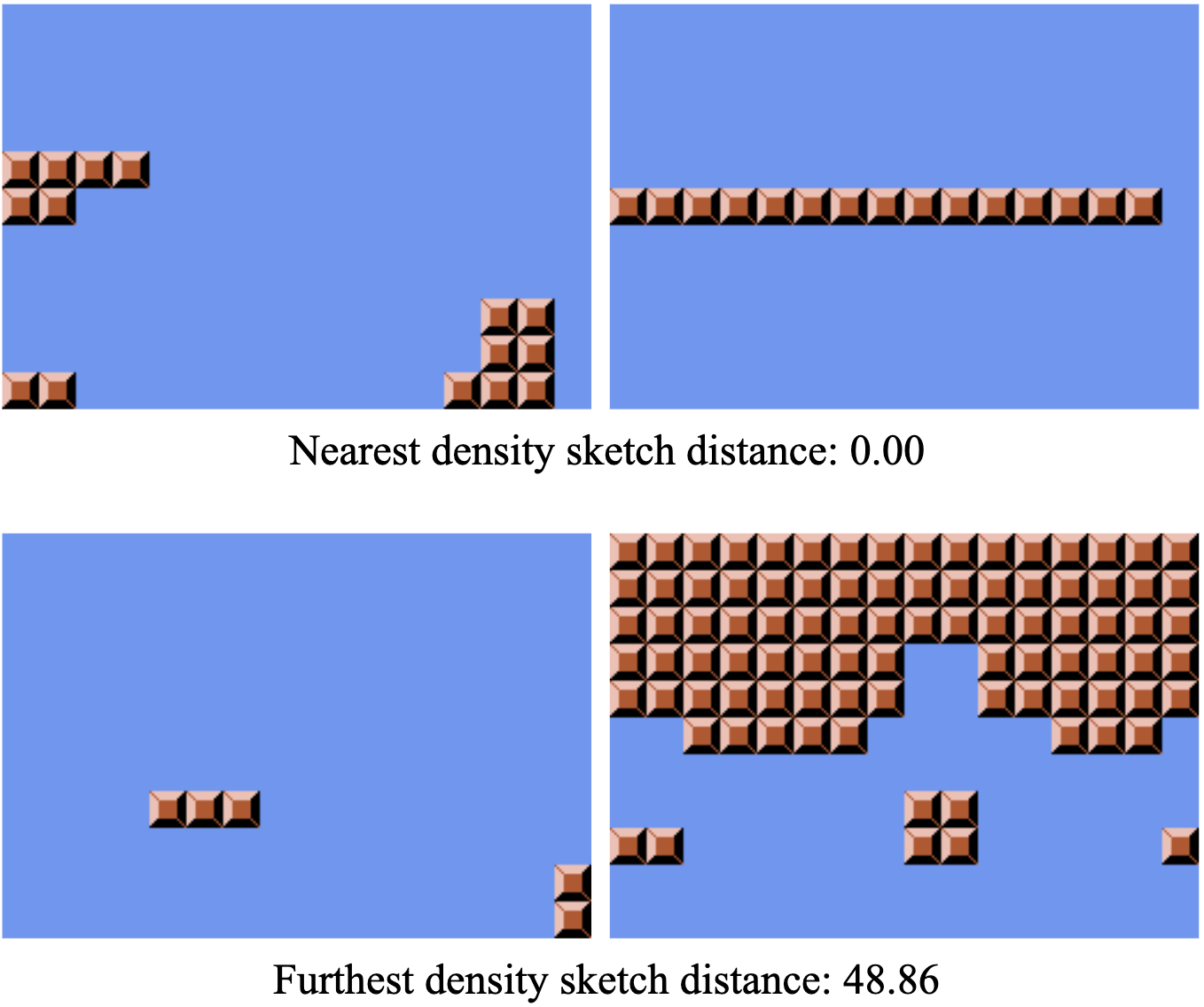} & \includegraphics[width=0.28\textwidth]{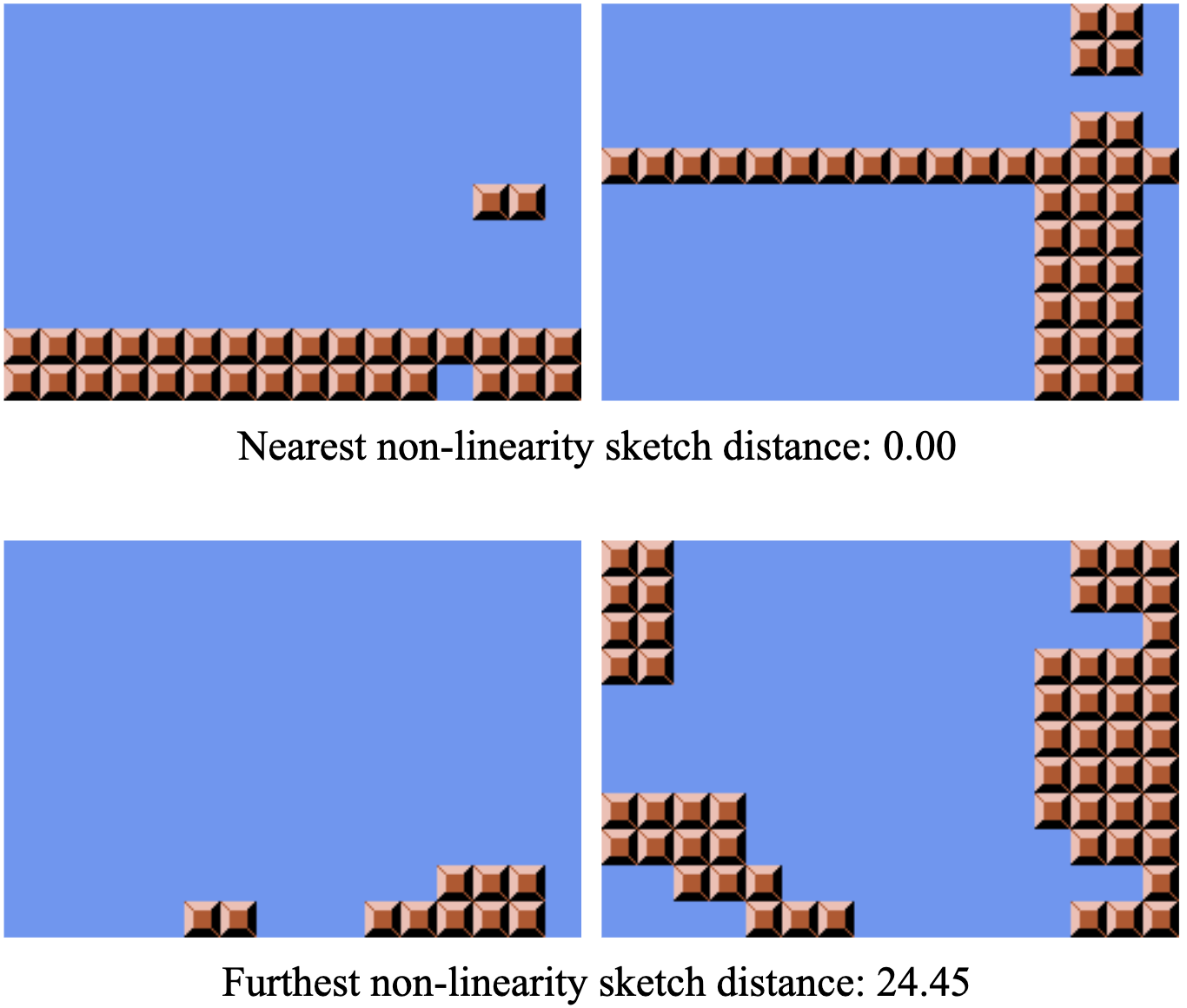} &
         \includegraphics[width=0.28\textwidth]{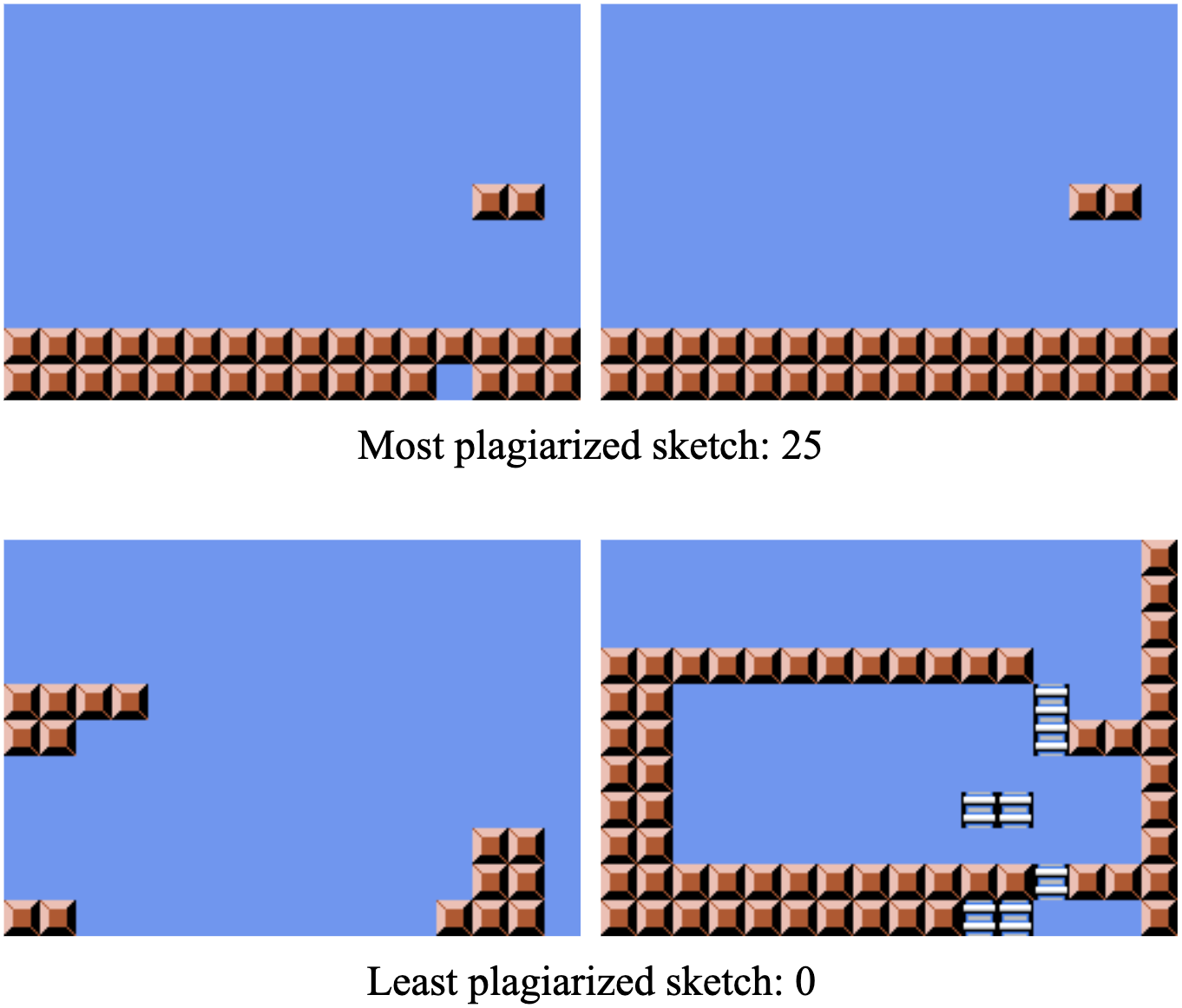}\\
   
        \multicolumn{3}{c}{(b) Comparison of generated \textit{LR} sketch sections and training sketch sections}\\
         \includegraphics[width=0.28\textwidth]{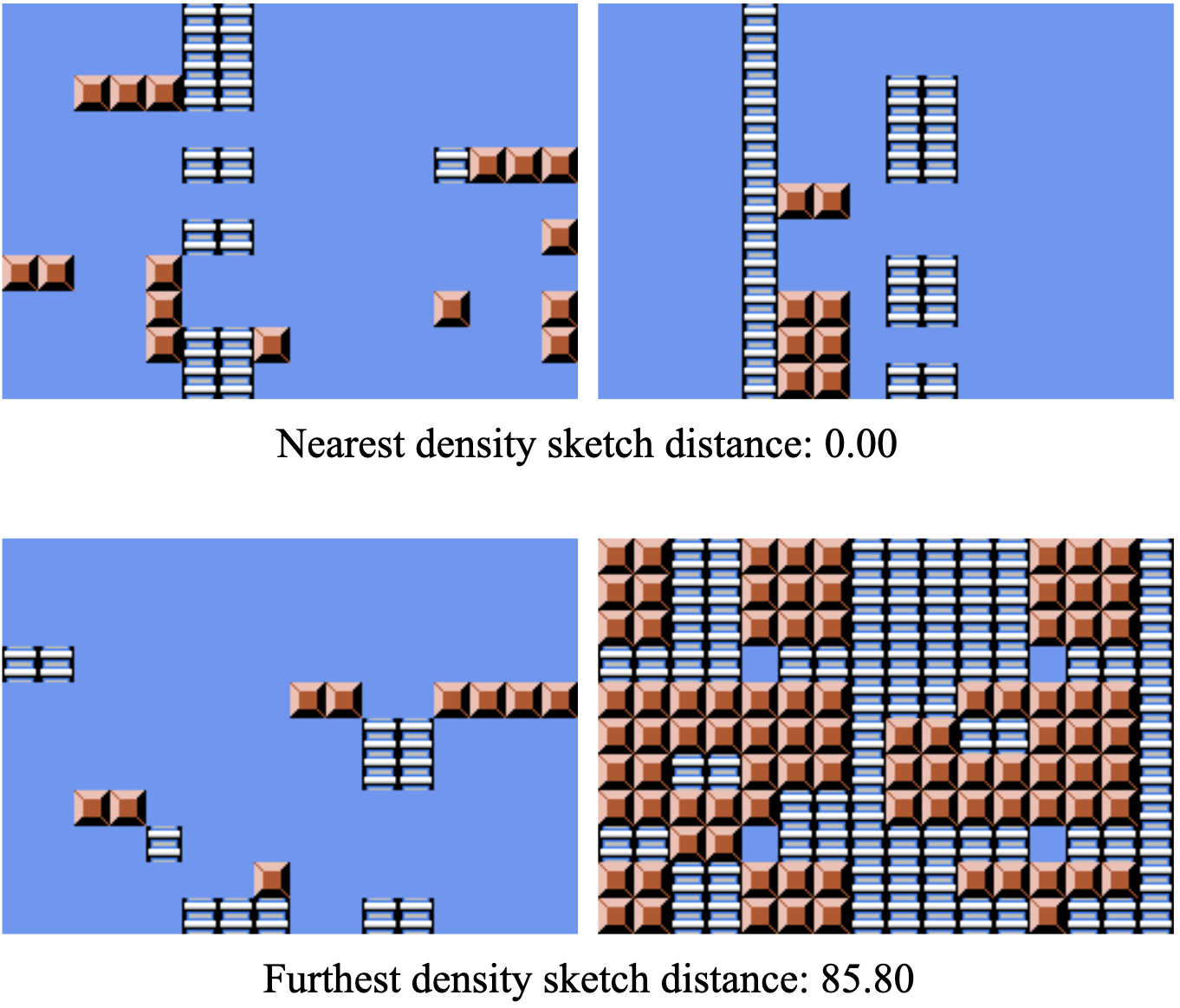} & \includegraphics[width=0.28\textwidth]{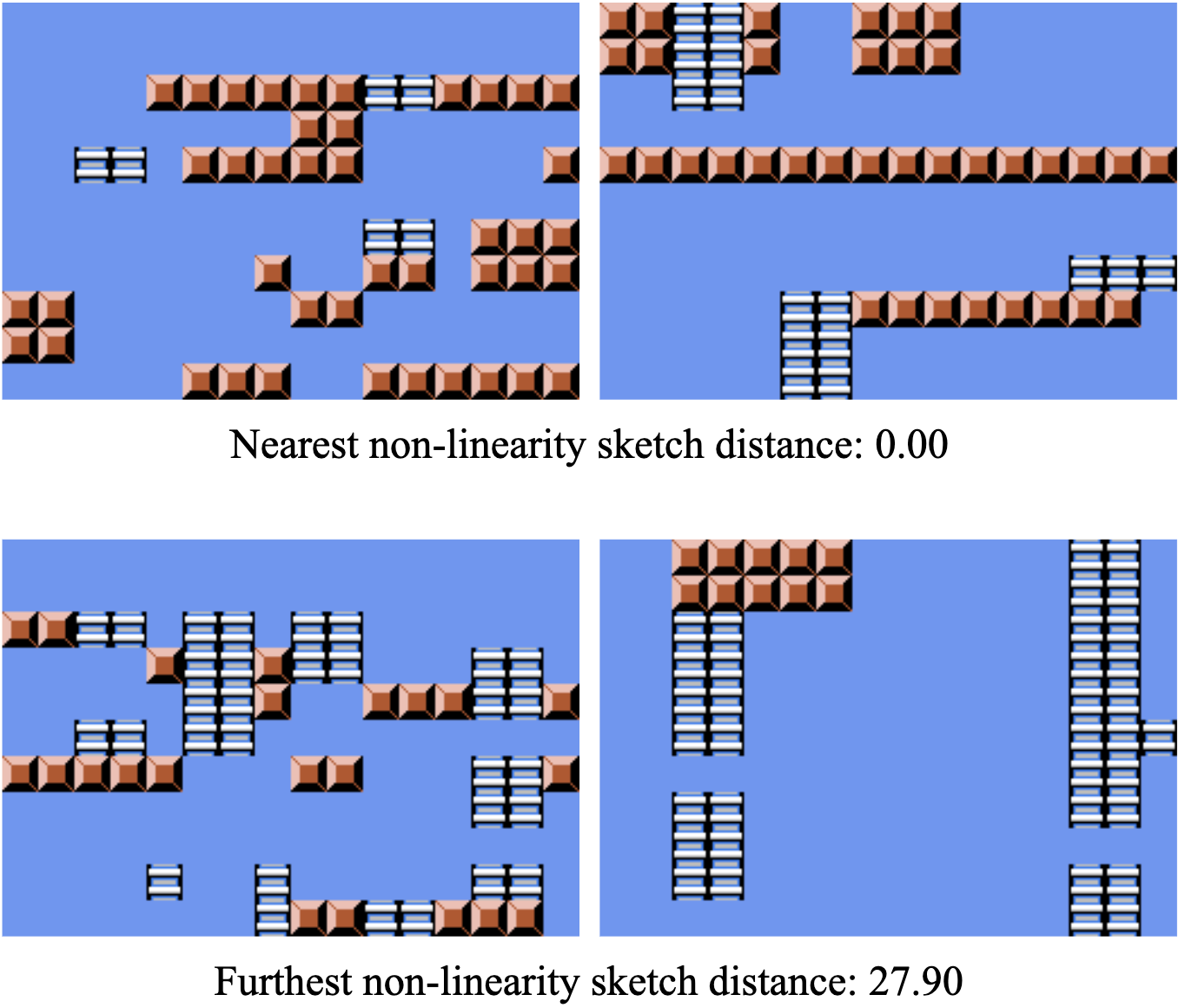} &
         \includegraphics[width=0.28\textwidth]{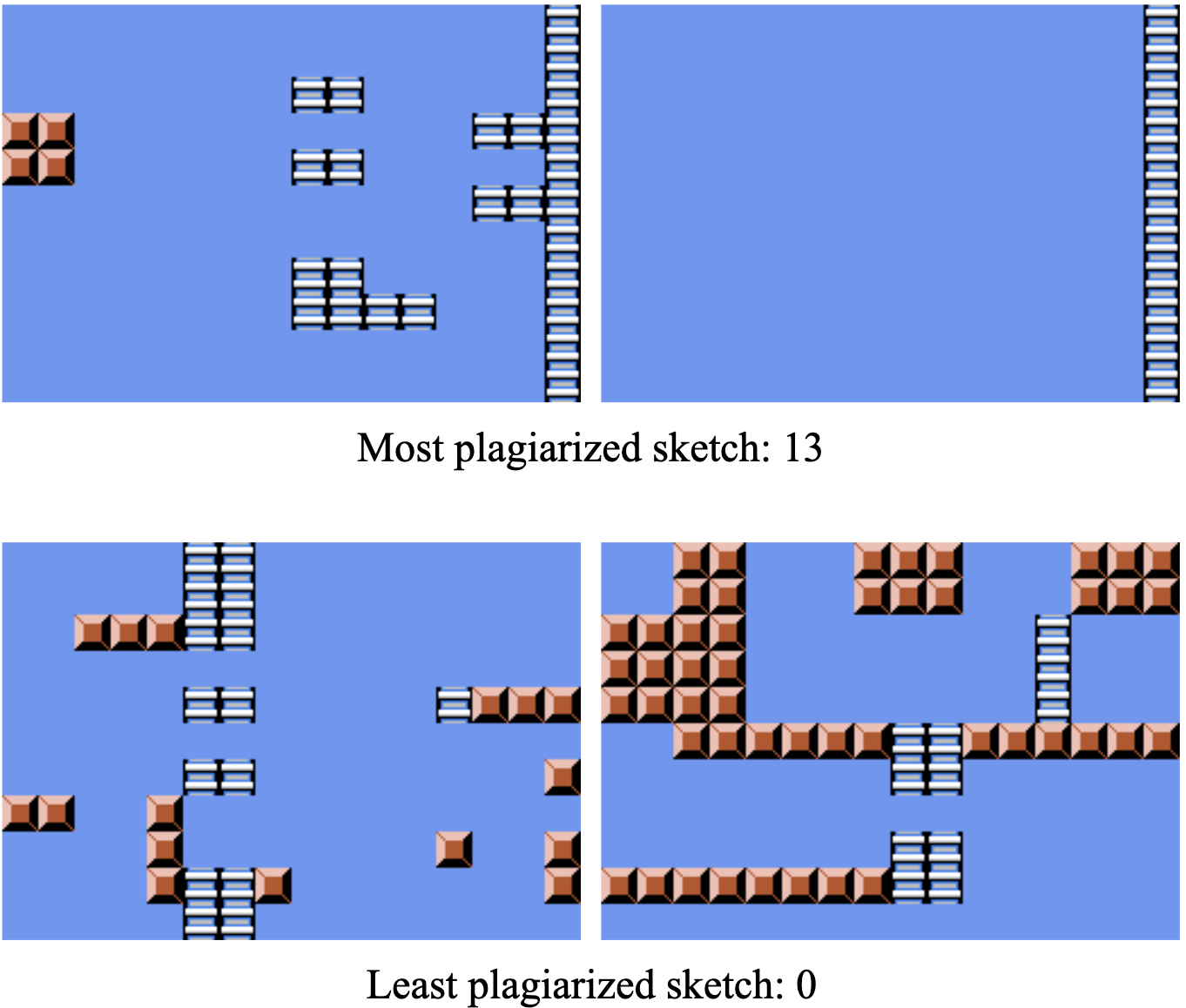}\\
   
        \multicolumn{3}{c}{(c) Comparison of generated \textit{MM} sketch sections and training sketch sections}\\
         \includegraphics[width=0.28\textwidth]{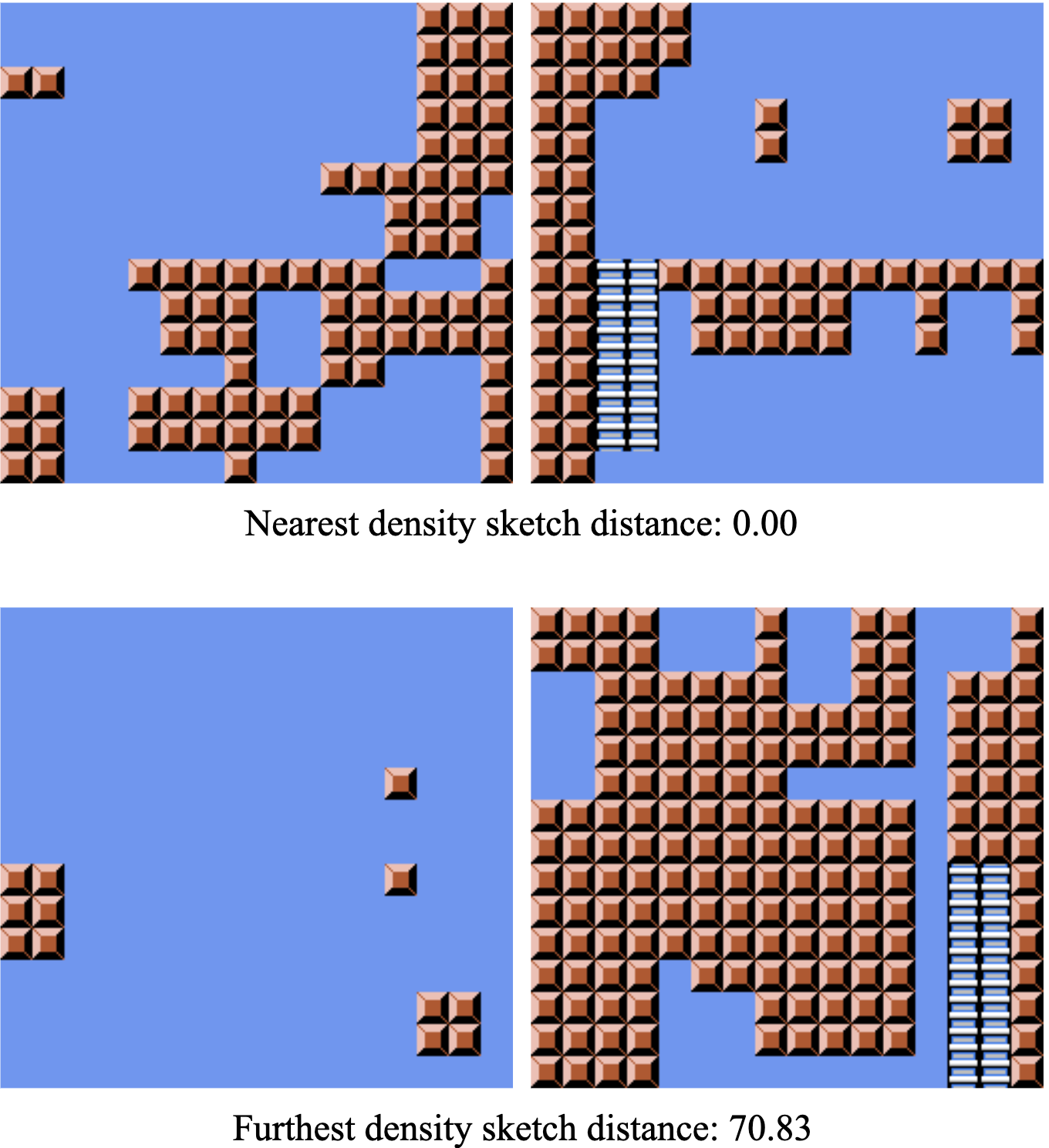} & \includegraphics[width=0.28\textwidth]{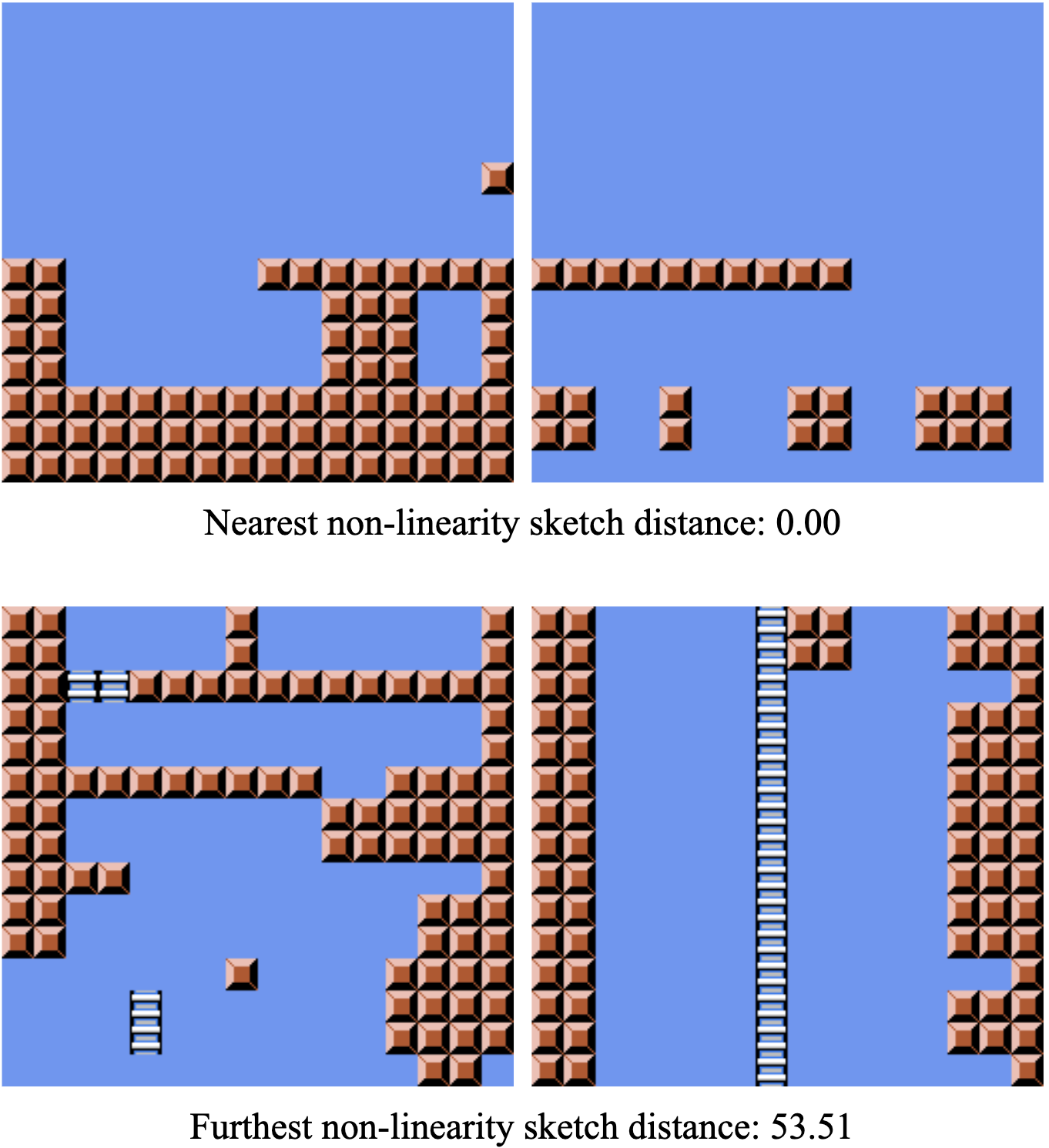} &
         \includegraphics[width=0.28\textwidth]{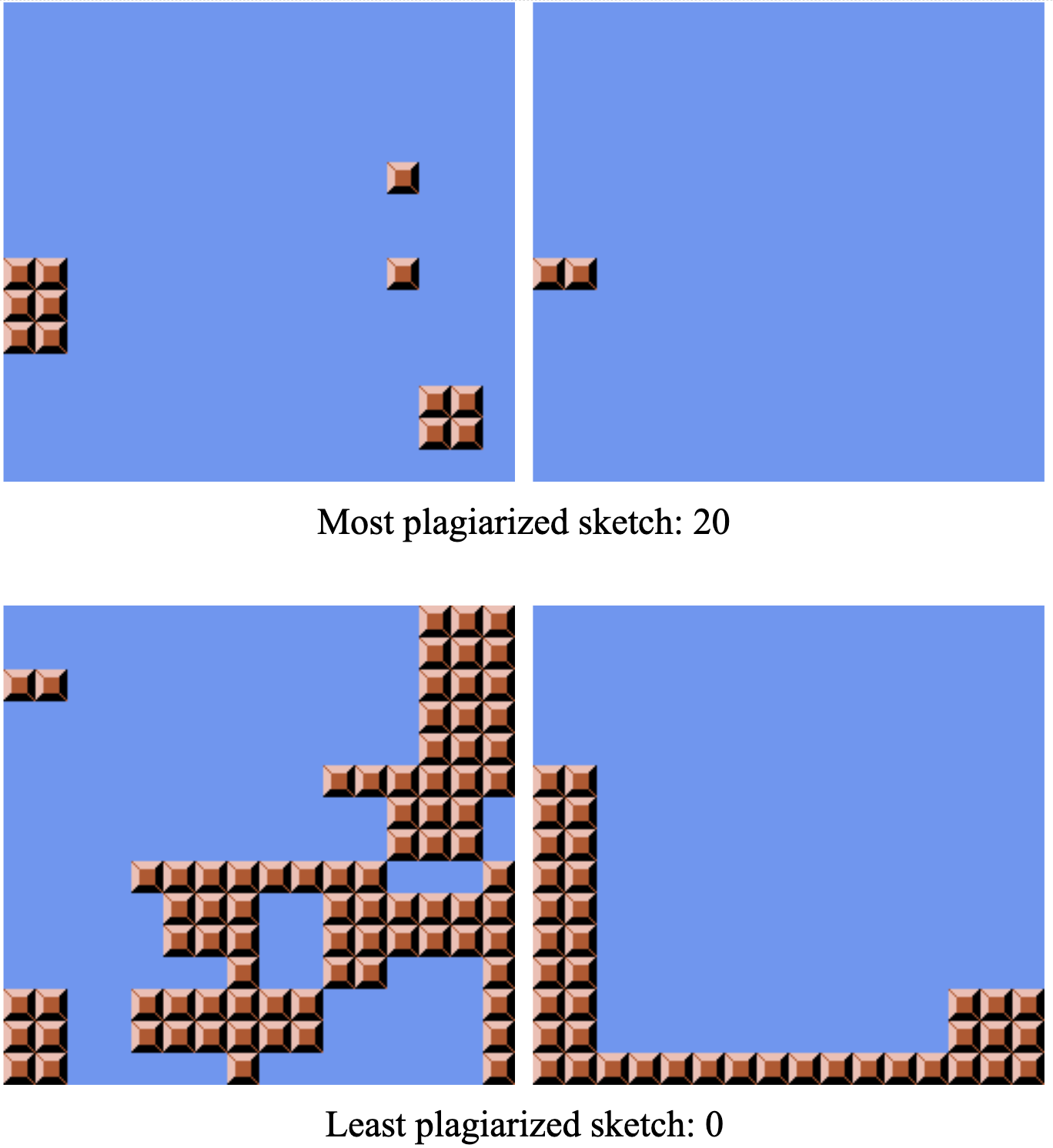}\\
    
        \multicolumn{3}{c}{(d) Comparison of generated \textit{NG} sketch sections and training sketch sections}\\
         \includegraphics[width=0.28\textwidth]{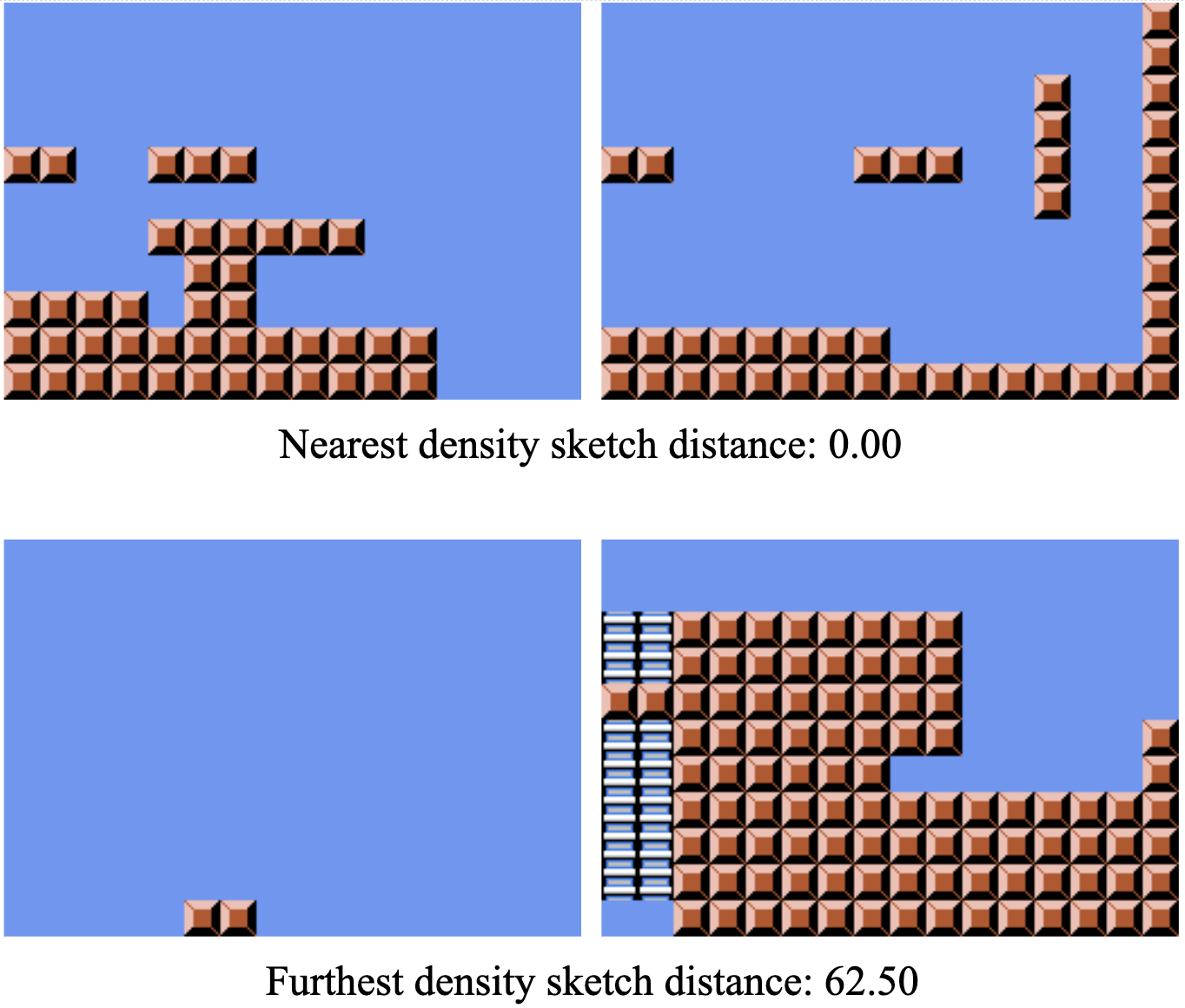} & \includegraphics[width=0.28\textwidth]{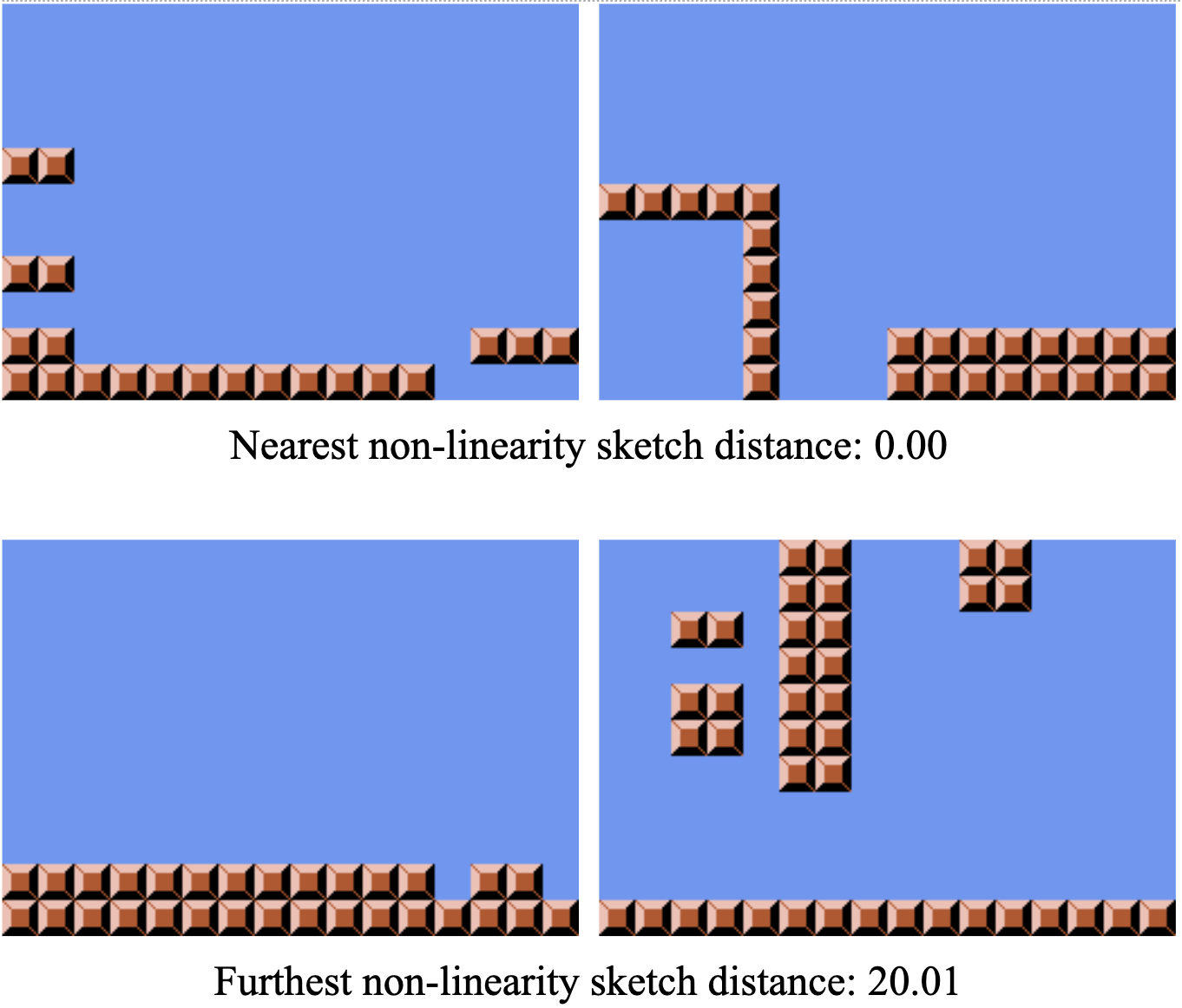} &
         \includegraphics[width=0.28\textwidth]{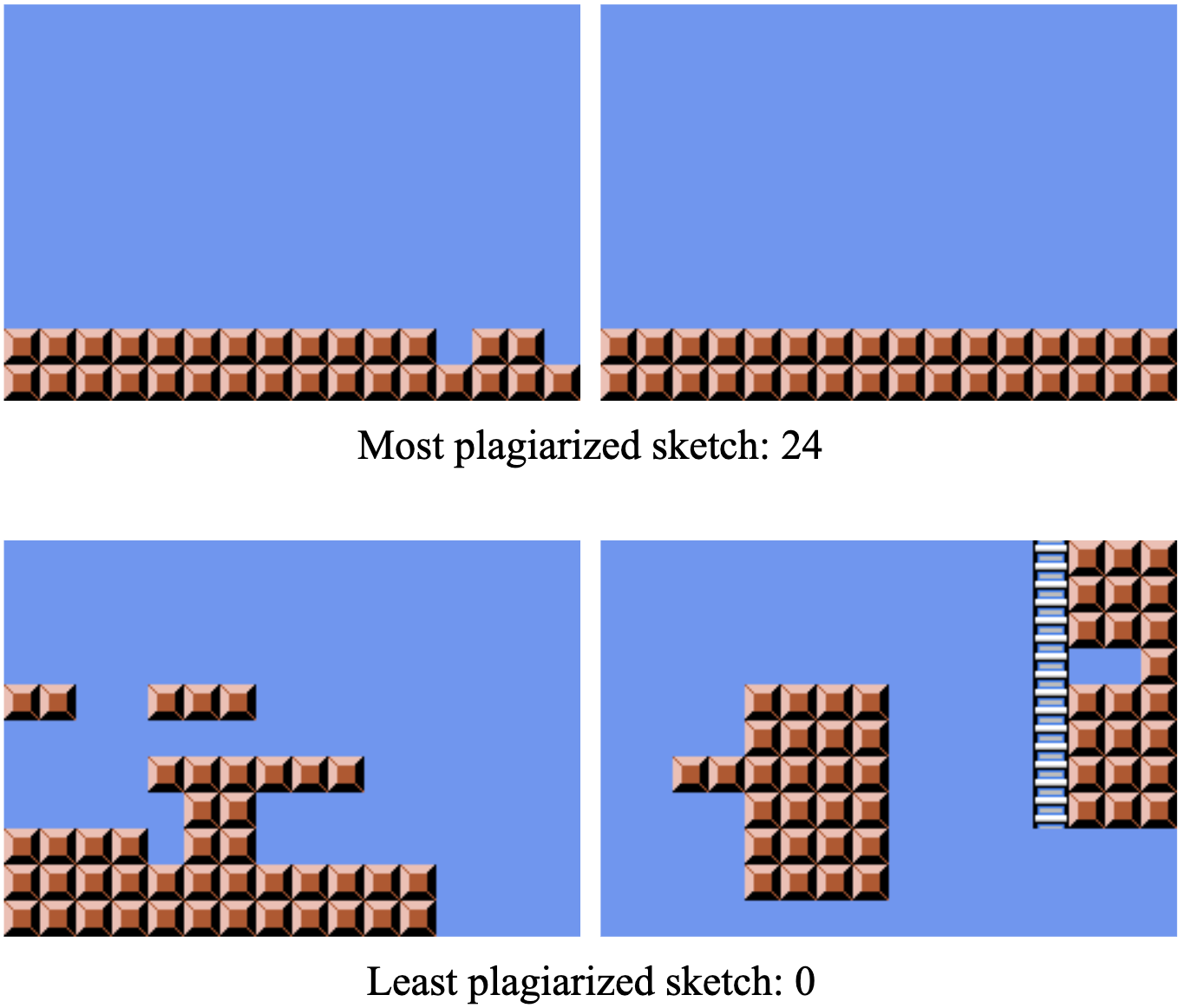}\\
        \end{tabular}
    \end{figure*}
    \begin{figure*}[h]
    \centering
        \begin{tabular}{ccc}
         \multicolumn{3}{c}{(e) Comparison of generated \textit{KI} sketch sections and training sketch sections}\\
         \includegraphics[width=0.28\textwidth]{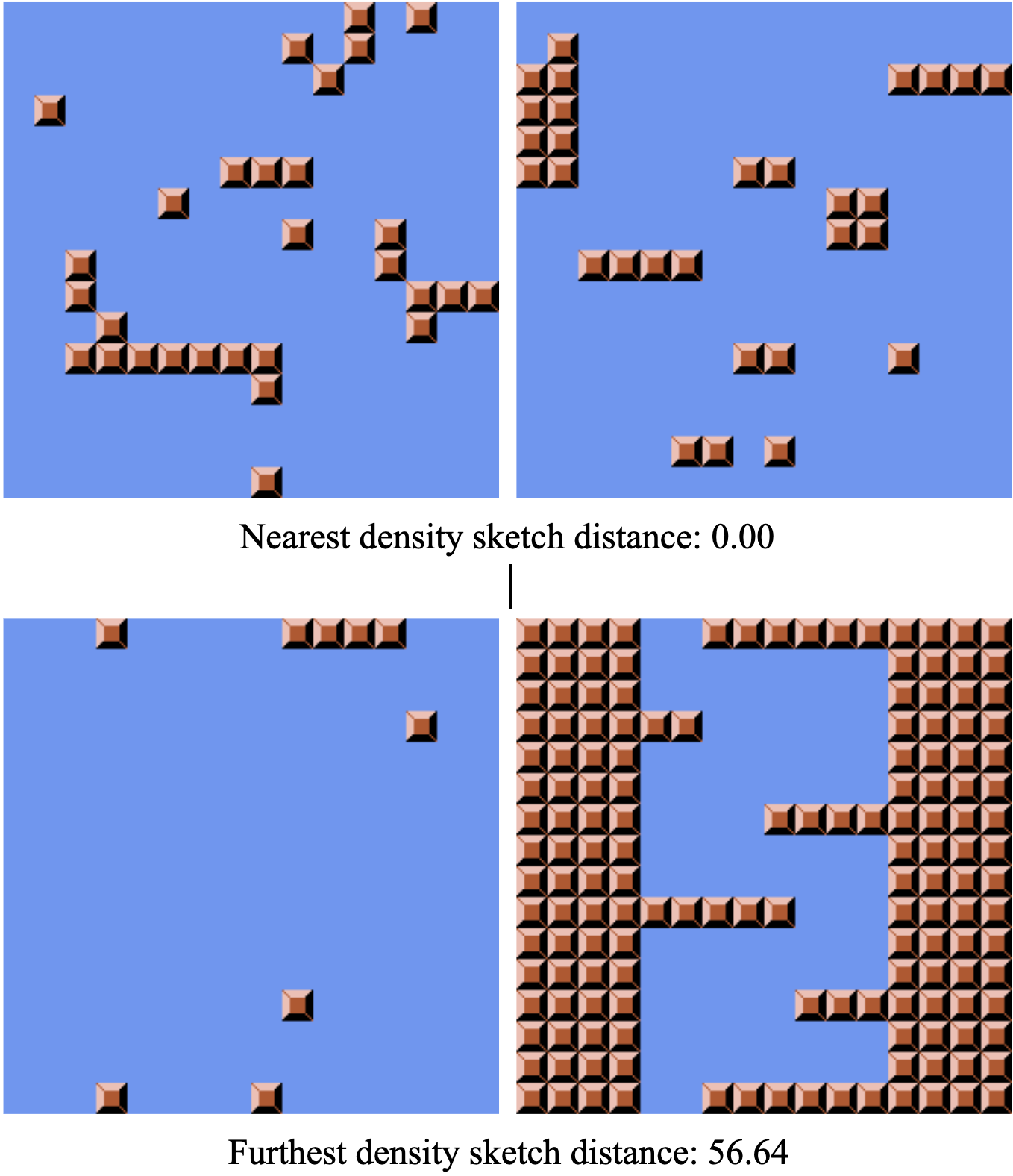} & \includegraphics[width=0.28\textwidth]{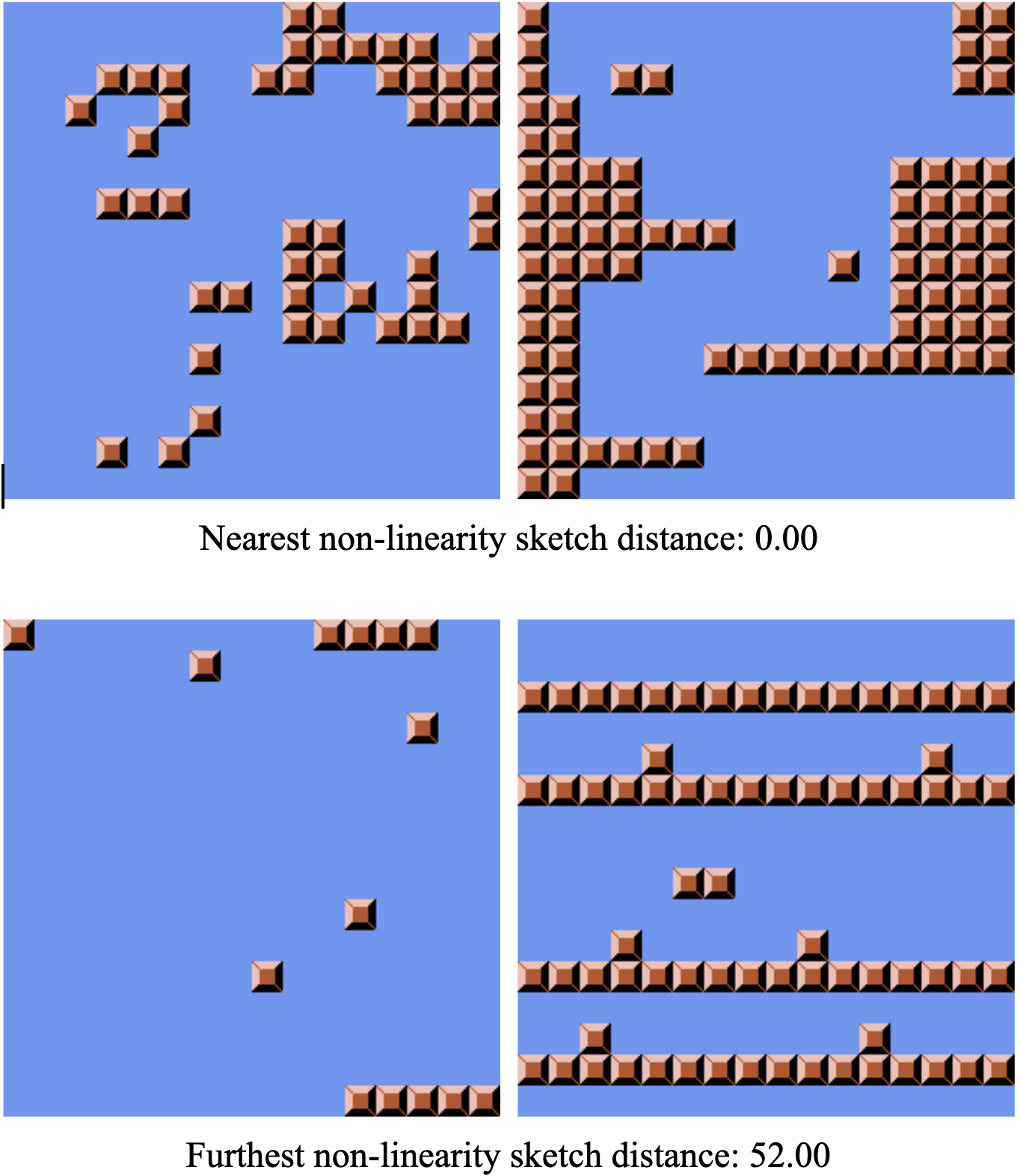} &
         \includegraphics[width=0.28\textwidth]{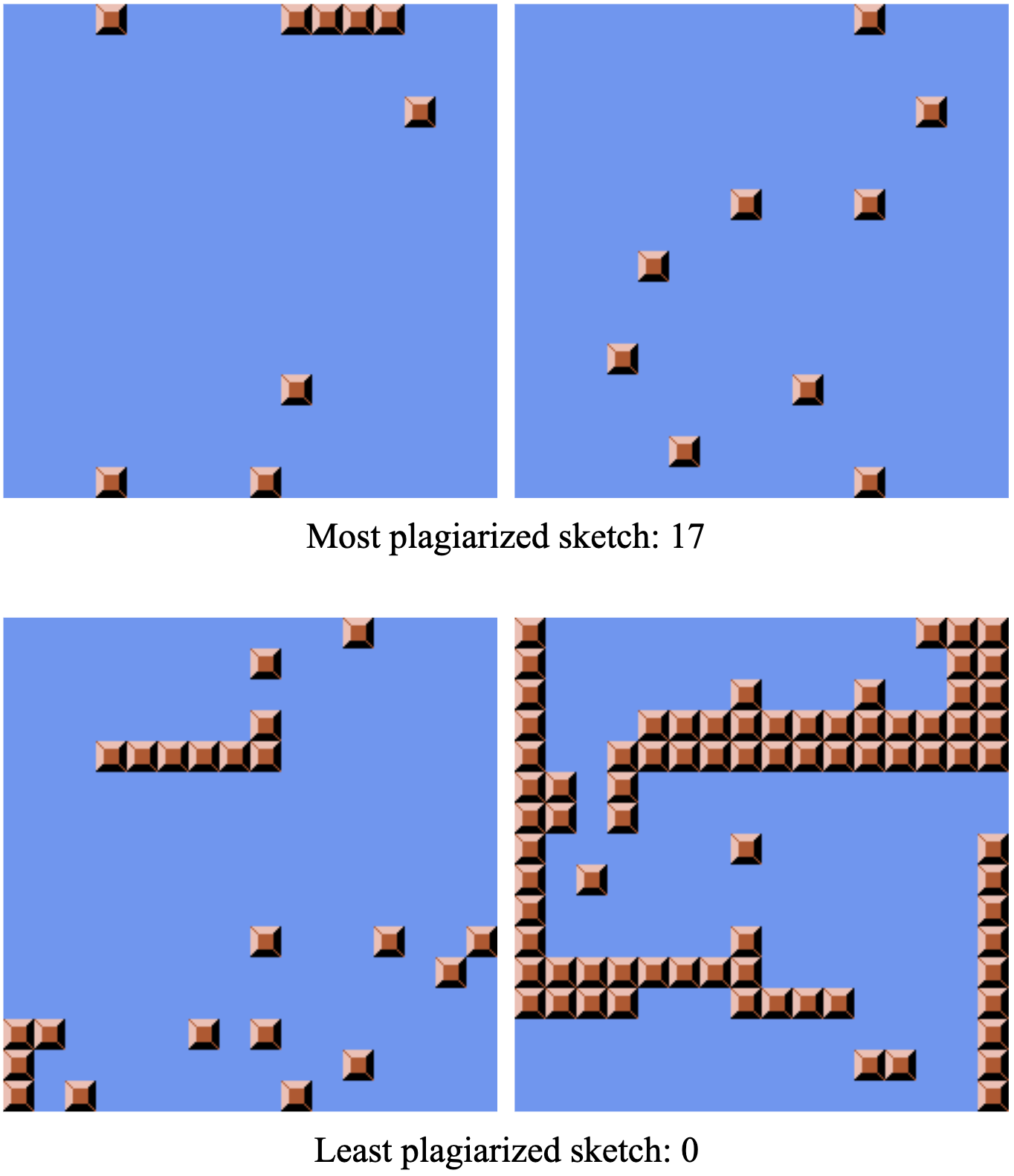}\\
    
        \multicolumn{3}{c}{(f) Comparison of generated \textit{MT} sketch sections and training sketch sections}\\
         \includegraphics[width=0.28\textwidth]{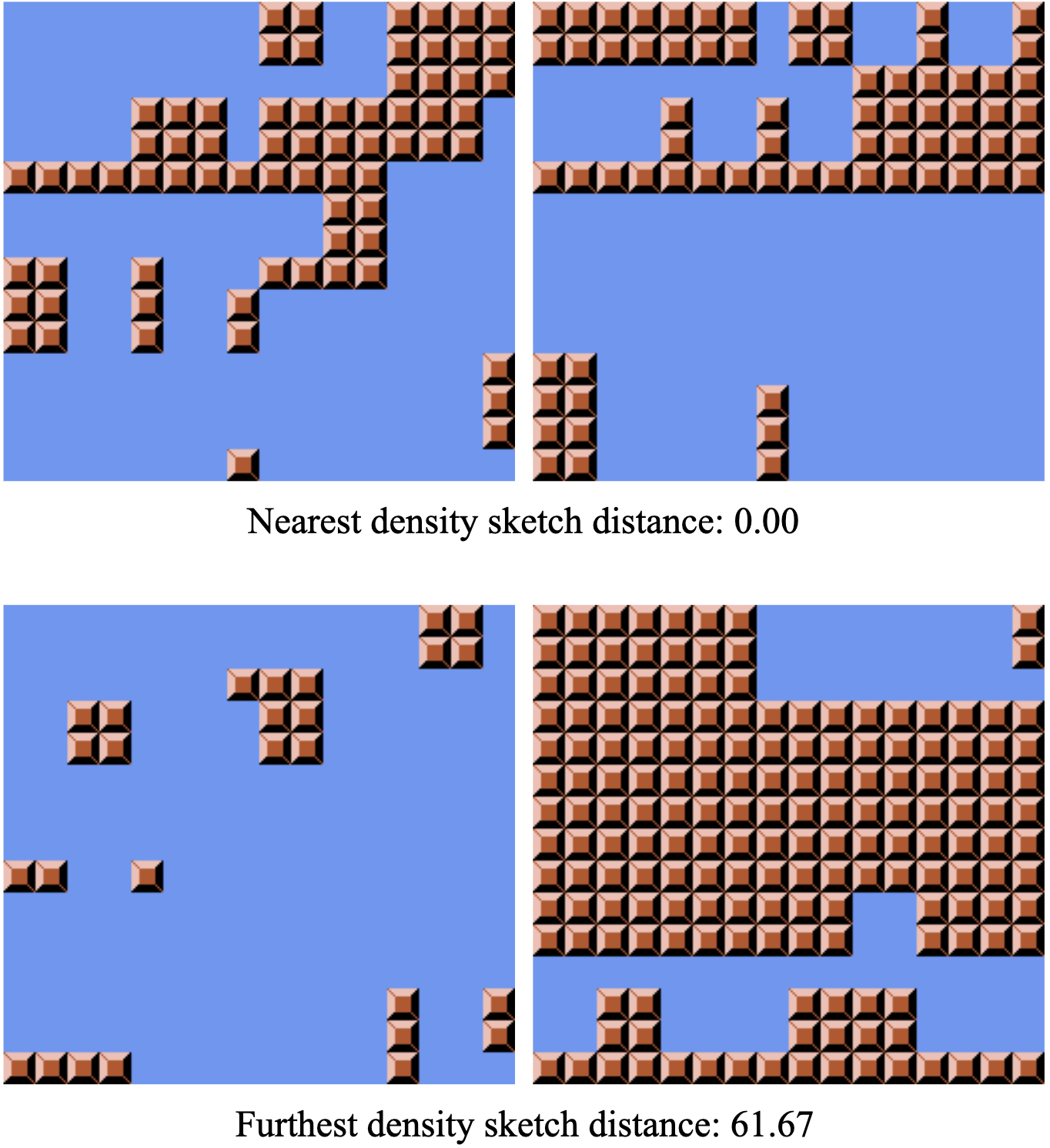} & \includegraphics[width=0.28\textwidth]{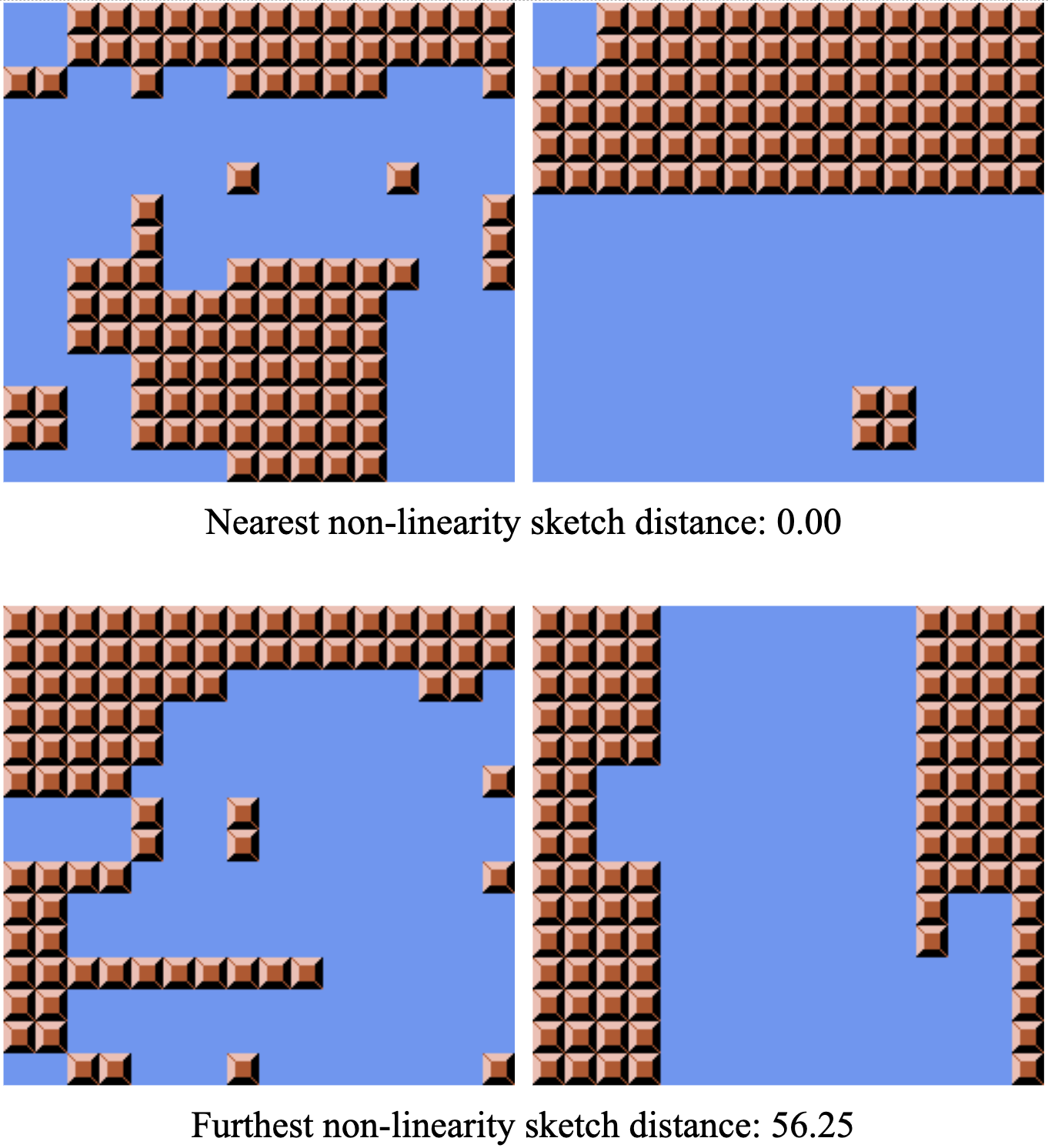} &
         \includegraphics[width=0.28\textwidth]{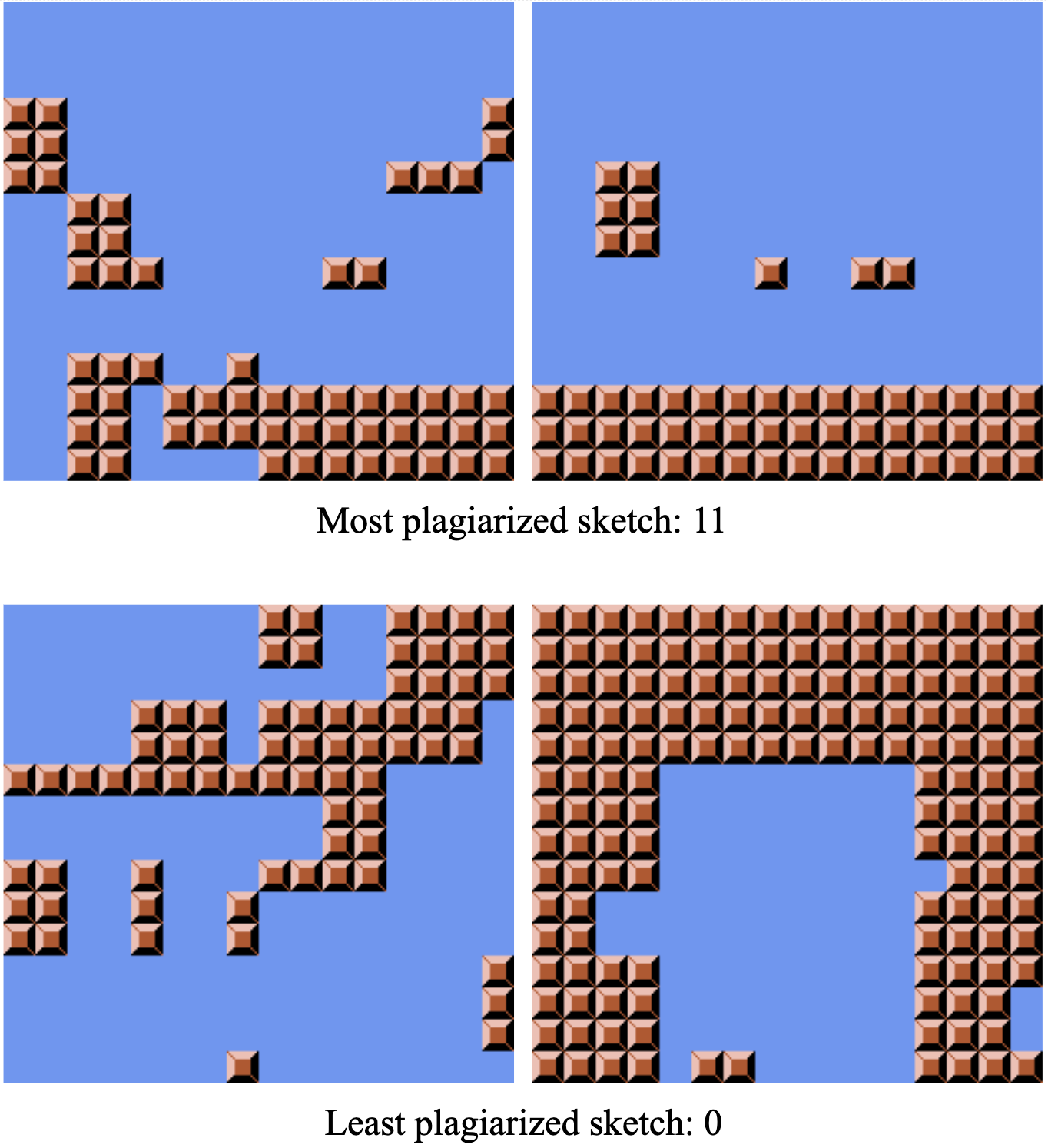}\\
    
        \multicolumn{3}{c}{(g) Comparison of generated \textit{SM} sketch sections and training sketch sections}\\
         \includegraphics[width=0.28\textwidth]{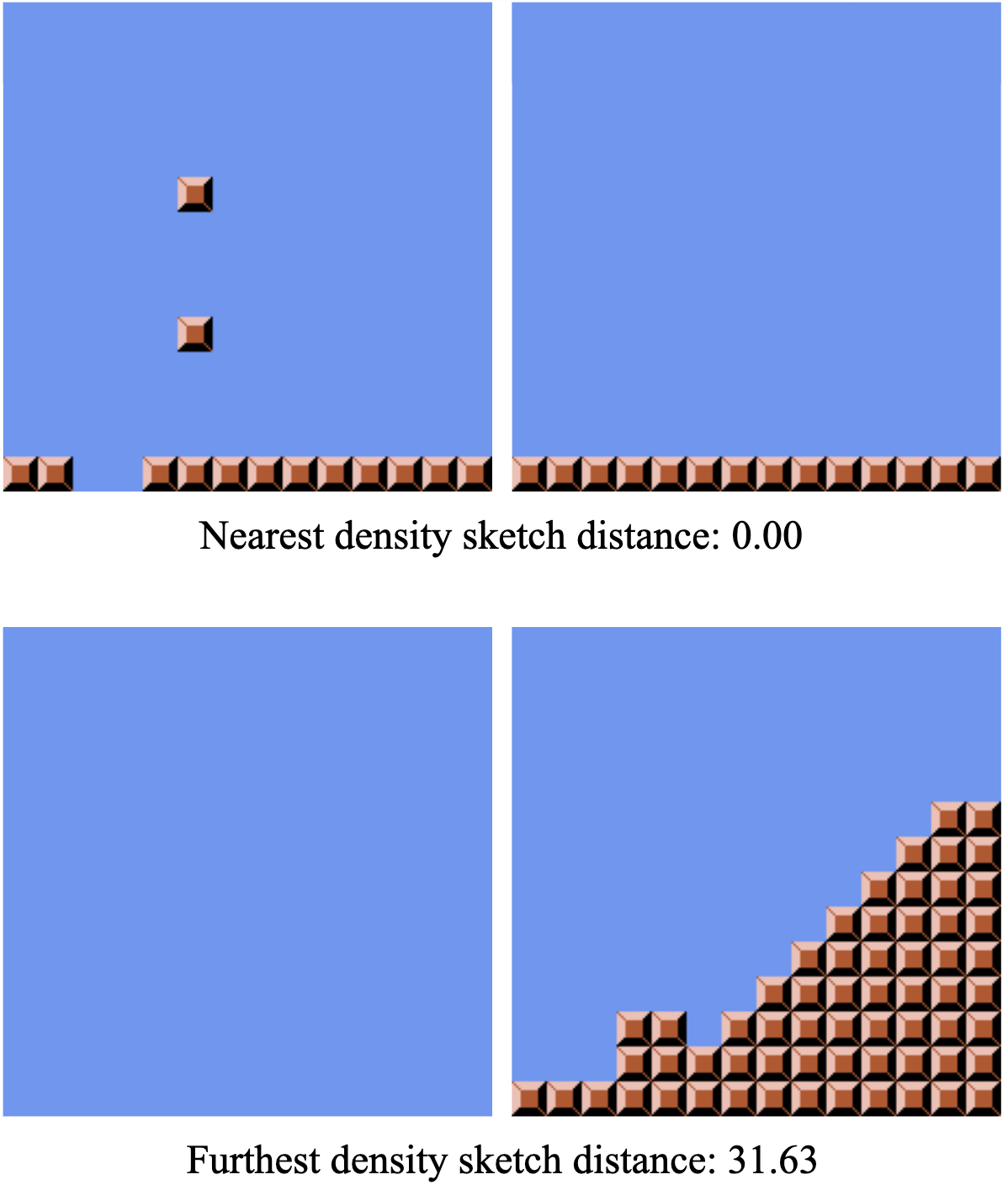} & \includegraphics[width=0.28\textwidth]{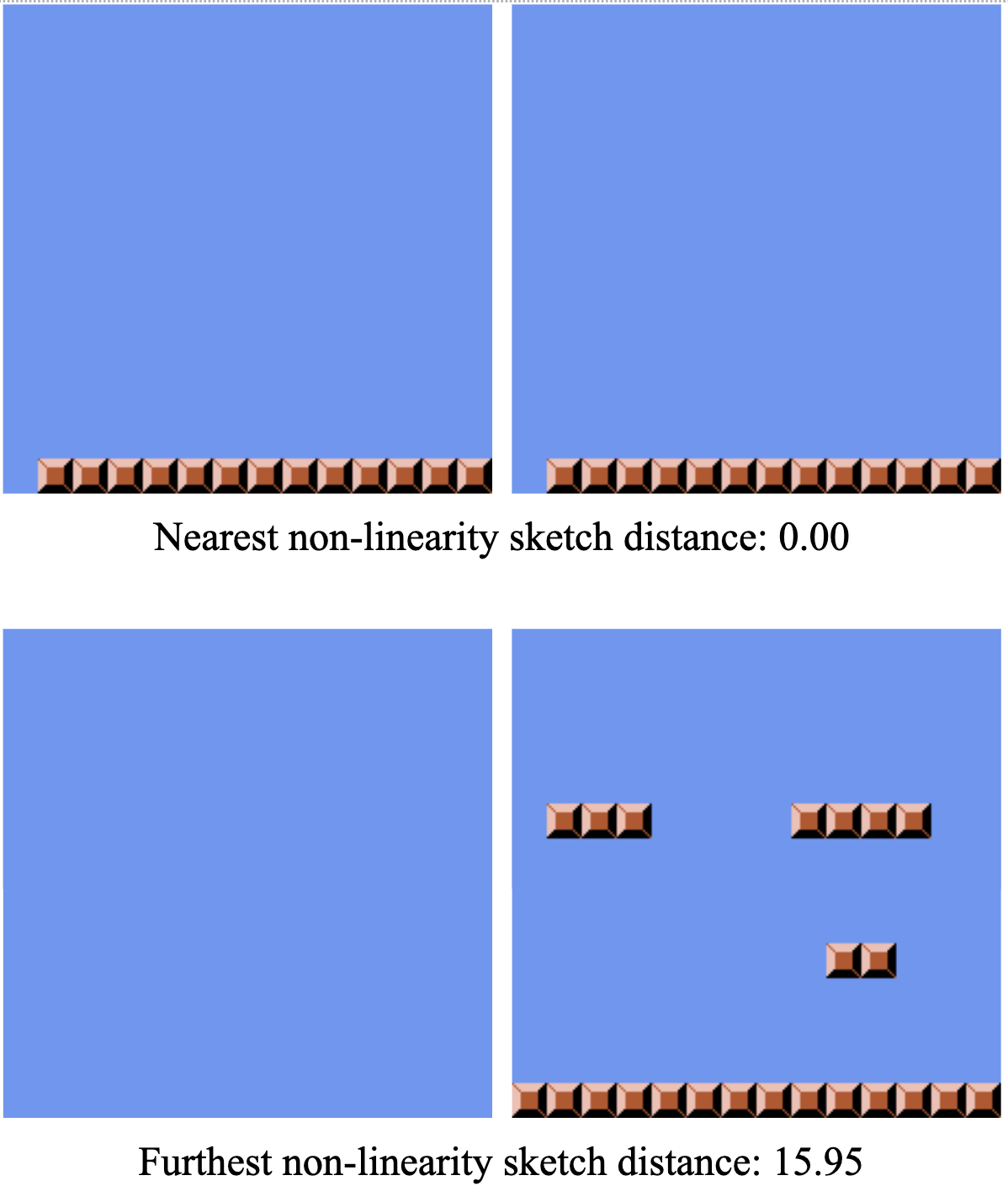} &
         \includegraphics[width=0.28\textwidth]{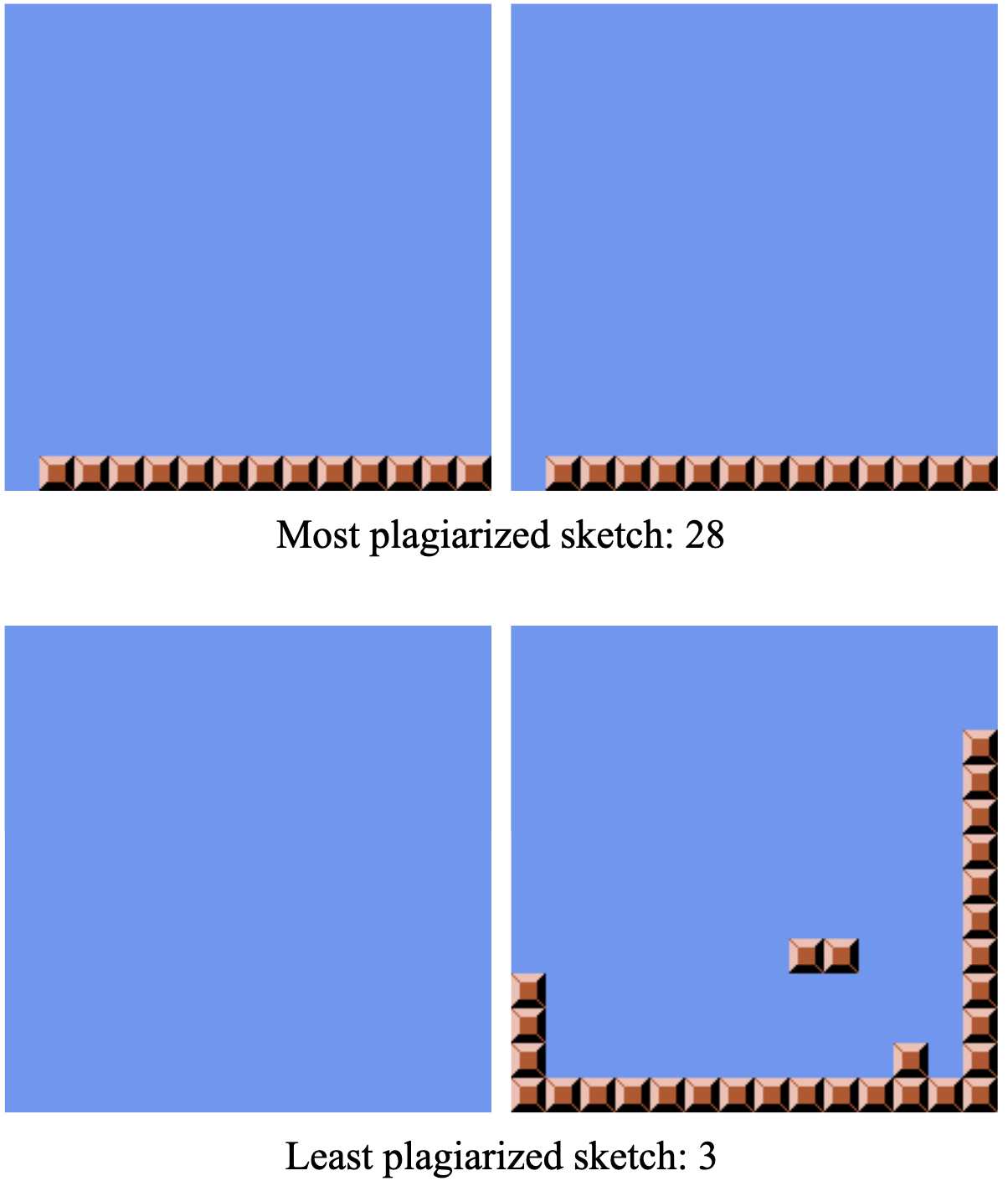}\\
    \end{tabular}
    \caption{This figure shows VAE-generated sketch sections for each domain compared with the nearest and furthest counterparts in the training levels, based on the evaluation metrics.}
    \label{fig:VAEcomp}
\end{figure*}

\clearpage
\newpage
\bibliographystyle{acm}
\bibliography{references}

\end{document}